\documentclass{article}

\PassOptionsToPackage{numbers,sort&compress}{natbib}
\usepackage[main,final]{neurips_2026}

\usepackage[utf8]{inputenc} 
\usepackage[T1]{fontenc}    
\usepackage{hyperref}
\hypersetup{hidelinks}       
\usepackage{url}            
\usepackage{booktabs}       
\usepackage{amsfonts}       
\usepackage{nicefrac}       
\usepackage{microtype}      
\usepackage{xcolor}         

\usepackage{times}
\usepackage{soul}
\usepackage{caption}
\usepackage{graphicx}
\usepackage{amsmath}
\usepackage{amsthm}
\usepackage{algorithm}
\usepackage{algorithmic}

\usepackage{wrapfig}
\usepackage{cleveref}
\usepackage{makecell}
\usepackage{multirow}
\usepackage{subcaption}
\usepackage[table,dvipsnames]{xcolor}
\usepackage{enumitem}
\newsavebox{\fullwidthtablebox}
\NewDocumentEnvironment{fullwidthtabular}{m m +b}{%
  \begingroup
  \sbox{\fullwidthtablebox}{\begin{tabular}{#2}#3\end{tabular}}%
  \setlength{\tabcolsep}{\dimexpr\tabcolsep+
    (\linewidth-\wd\fullwidthtablebox)/\numexpr2*#1\relax\relax}%
  \begin{tabular}{#2}#3\end{tabular}%
  \endgroup
}{}

\title{BiHashFormer: Hash-Driven Dual-Branch Transformer for Efficient Object Detection in HRW Shots}

\author{%
  Jingchen Huang\thanks{Equal contribution.} \\
  Department of Automation \\
  Tsinghua University \\
  Beijing, China \\
  \texttt{huang-jc23@} \\
  \texttt{mails.tsinghua.edu.cn}
  \And
  Wenxi Li\footnotemark[1] \\
  KLATASDS-MOE \\
  School of Statistics \\
  East China Normal University \\
  Shanghai, China \\
  \texttt{wxli@sfs.ecnu.edu.cn}
  \And
  Chenyang Lyu\footnotemark[1] \\
  Alibaba Token Hub \\
  Alibaba Group \\
  Hangzhou, China \\
  \texttt{lyuchenyang.dcu@} \\
  \texttt{gmail.com}
  \AND
  Moran Liu \\
  Beijing National Research \\
  Center for Information \\
  Science and Technology \\
  Department of Automation \\
  Tsinghua University \\
  Beijing, China \\
  Zhuoxi Lab, Hangzhou, China \\
  \texttt{liumoreran@gmail.com}
  \And
  Haozhe Lin\thanks{Corresponding authors.} \\
  Beijing National Research \\
  Center for Information \\
  Science and Technology \\
  Tsinghua University \\
  Beijing, China \\
  \texttt{linhz@tsinghua.edu.cn}
  \And
  Yuchen Guo\footnotemark[2] \\
  Beijing National Research \\
  Center for Information \\
  Science and Technology \\
  Tsinghua University \\
  Beijing, China \\
  \texttt{yuchen.w.guo@} \\
  \texttt{gmail.com}
}

\begin{document}

\setcounter{footnote}{0}
\maketitle
\setcounter{footnote}{0}

\begin{abstract}
Recent advances in gigapixel-level imaging have brought High-Resolution Wide shots to the forefront of research. However, these images present significant challenges: extreme sparsity of foreground, gigapixel-level resolutions and interleaving of foreground and background. This causes traditional detectors and attention mechanisms to be hindered by background, resulting in inefficiency and inaccuracy.
To tackle this problem, we propose BiHashFormer, a dual sparsified branch transformer built on hashing. RoI Selector will predict the foreground proportion for top-k selecting target-containing windows. The compression branch processes windows selected and refines attention into HashAttention by discarding half of the key–value pairs, reducing cost and improving efficiency. The compensation branch uses HashMiner to perform low-cost hash searches in the remaining regions, recovering features of missed objects.
Experiments confirm that the selection relationship in HashAttention is one-way and reveal the preference of different queries for keys. In experiments on the gigapixel benchmark PANDA, BiHashFormer reduces 37.5\% backbone FLOPs while improving $\text{AP}_{50}$ to 81.0\%. Moreover, this parallel dual-branch structure allows the model to avoid inference delays caused by inter-branch dependencies, while achieving much greater stability than single-branch models, specially with low rates of RoI selection.
\end{abstract}

\section{Introduction}
In computer vision, object detection has long been recognized as a central challenge, requiring algorithms to simultaneously determine the location and category of target instances within an image. The standard output of such systems usually consists of a bounding box together with a class label. Landmark approaches, exemplified by Fast R-CNN \cite{girshick2015fast}, have achieved notable success on large-scale benchmarks such as MS COCO \cite{lin2014microsoft}. More recently, advances in nondestructive imaging have shifted attention toward High-Resolution Wide (HRW) detection \cite{chen2022towards,fan2022speed}. This setting involves analyzing imagery with both extremely fine spatial detail and expansive coverage, posing unique computational and modeling challenges. Representative scenarios include gigapixel drone photography \cite{svanstrom2021real}, satellite-based observation \cite{DOTA,ding2021object,ding2018learning}, and panoramic camera data \cite{360+x}.

\begin{wrapfigure}{r}{0.5\linewidth}
    \centering
    \includegraphics[width=1\linewidth]{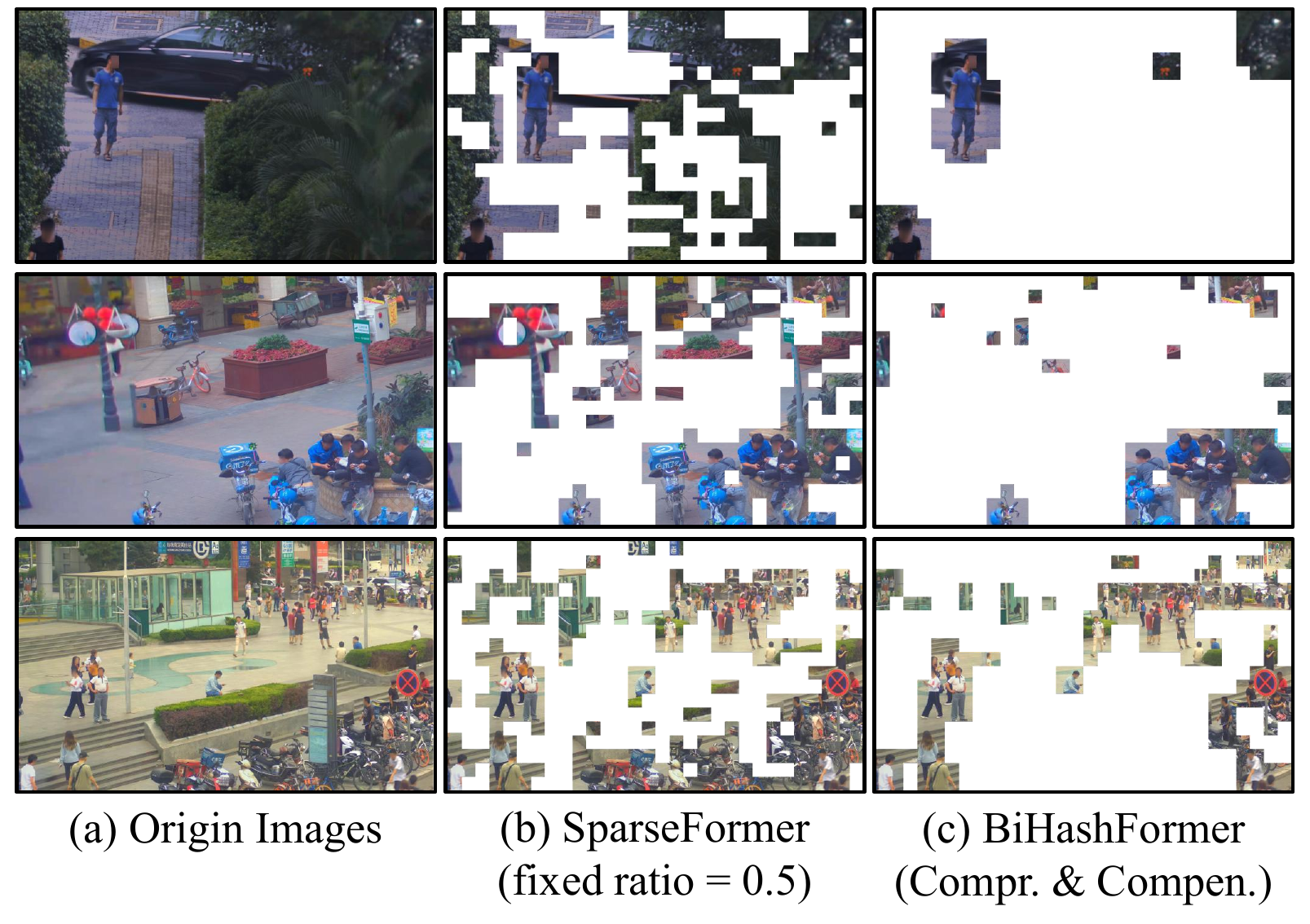}
    \caption{\textbf{Comparison of RoI between different backbones.} (b) SparseFormer selects a fixed ratio of regions for processing, regardless of input variability. (c) BiHashFormer combines lower selection ratio in the compression branch with dynamic compensation, making it more flexible and efficient. For the compensation branch, windows are drawn if any key in them is selected.}
    \label{fig:1}
\end{wrapfigure}

The first problem in HRW detection is that the objects we care about only take up a very small part of the whole picture. In normal detection datasets, the foreground can cover more than half of the image. But in HRW scenes, the foreground is usually less than 15\%~\cite{SparseFormer}, because the images come from large real-world views that naturally include wide areas of background such as sky, buildings, or trees. This means detectors have to search through a lot of background to find a few targets, and even strong models like DINO \cite{DINO}, Deformable DETR \cite{DETR}, and RMT \cite{RMT} often make mistakes when the background is too dominant. This problem is illustrated by \Cref{fig:1}.

The second challenge in HRW detection is the heavy demand on computation. Many application scenarios, such as UAV inspection (VisDrone \cite{VisDrone}) or medical analysis (VL-Health \cite{VL-Health}), require detectors to respond in real time. Yet as image resolution grows—from millions of pixels in COCO \cite{lin2014microsoft} to billions in PANDA \cite{PANDA}—the processing cost rises sharply. Conventional detectors scale poorly with input size, and inference speed can drop by nearly two orders of magnitude, making real-time deployment difficult.

Thirdly, the strong interleaving of foreground and background. HRW detection often requires larger windows (in Swin \cite{Swin}) to capture sufficient context, especially in crowded scenes. Yet enlarging the window in conventional attention introduces excessive background tokens, diluting the focus on true targets. For example, a person on a lawn may be overshadowed because the cumulative attention to numerous lawn tokens exceeds that of the human body, causing facial tokens to misdirect attention and lowering detection accuracy. Furthermore, this makes small objects or those surrounded by background more likely to be excluded during RoI selection, as shown in \Cref{fig:1}.

Inspired by the widespread use of hashing in computer science, we propose BiHashFormer which combines dual-branch design with hashing, achieving both sparse attention and robust compensation for overlooked targets. Firstly, a variance-based scoring mechanism ranks windows, and a learned foreground predictor determines the image-dependent proportion of RoIs. For the selected RoIs, a sparse Attention mechanism called HashAttention is introduced. This approach compresses the key-value pairs for each query, leading to finer-grained computational reduction. At the same time, the quality of attention within RoIs will be improved. In the remaining regions, detection targets are localized within the Hamming feature space obtained through hash projection. Prototype encoding is employed for this localization and such compensation can be achieved with extremely low computational cost. On PANDA \cite{PANDA}, BiHashFormer achieves a 37.5\% reduction in FLOPs compared with SparseFormer \cite{SparseFormer}, while significantly improving $\text{AP}_{50}$.

In summary, our key contributions are:

\begin{itemize}
\item We propose BiHashFormer, a dual-branch vision backbone based on transformer, which combines fine-grained sparse computation with global target compensation. It is able to effectively handle sparsely distributed objects intertwined with background in HRW images while reducing computational cost.

\item We introduce HashAttention for the compression branch. By projecting features into a low-dimensional Hamming space, similarity computation is greatly reduced. The model compresses interactions between query and key-value pairs, achieving RoI-level sparse computation and refining attention results.

\item We design HashMiner for the compensation branch. In the Hamming feature space, the model learns prototype codes for detection targets, enabling retrieval of objects from non-RoI regions with minimal overhead and providing compensation for small or difficult targets.

\item We develop a label-free strategy with a special loss for prototype codes. Leveraging information from the compression branch, the model learns codes that capture broader and more accurate human-related semantics.

\end{itemize}

\section{Related work}
\textbf{Object detection in high-resolution wide shots.}
High-resolution wide-shot (HRW) detection focuses on the challenge of identifying sparse targets within gigapixel-scale images. Unlike traditional benchmarks such as ImageNet \cite{Imagenet}, HRW scenes are characterized by extremely low target density (often below 15\% coverage) and highly diverse backgrounds, as highlighted in PANDA \cite{PANDA}. Although mainstream detectors, including Faster R-CNN \cite{girshick2015fast} and Deformable Convolutional Networks \cite{DeformableCN}, perform well in dense environments, their computational burden grows rapidly with image resolution, leading to unacceptable latency. Shufflenet \cite{Shufflenet} improves efficiency by employing grouped convolutions to reduce complexity, yet it lacks mechanisms for selective computation across image regions. DynamicDet \cite{Dynamicdet} introduces adaptive inference by routing inputs through different pathways based on estimated difficulty, but its binary partitioning strategy is unsuitable for HRW scenarios and the dual-branch design adds unnecessary overhead. Similarly, DGE \cite{DGE} has been utilized to strengthen vision transformers, but its rigidity limits adaptability to the unique demands of HRW detection.

\noindent\textbf{Vision transformers for detection.}
Transformers have achieved remarkable success in the field of natural language processing (NLP) and Vision Transformers (ViTs) are likewise revolutionizing detection by modeling long-range dependencies through attention. Swin \cite{Swin} mitigates the rising computational cost of large images with window-based attention, yet falls short of the contextual demands of HRW images. Dual-branch models like DBCT-Net \cite{wang2023dbct} fuse global and local features but fail to extract task-relevant information from massive non-RoI regions, leading to inefficiency in HRW detection. They share a common limitation: either treat all tokens equally or focus only on selected RoIs, overlooking the small and hard targets that are prevalent in HRW scenarios.

\noindent\textbf{Sparse attention methods.}
To enhance the effectiveness of attention mechanisms in HRW detection, numerous sparsification approaches have been proposed. XAttention \cite{xu2025xattention} leverages the sum along the anti-diagonal of the attention matrix to selectively compute more important blocks, which binds queries and key–value pairs together and limits expressiveness. Linformer \cite{wang2020linformer} applies low-rank projections to compute attention in reduced dimensions, trading semantic richness for lower cost. DSA \cite{liu2022dynamic} attempts top‑k sparsification based on attention weights from q–k dot products, but the computational overhead of the dot product itself is prohibitive in HRW scenarios.

\begin{figure*}[t]
    \centering
    \includegraphics[width=1\textwidth]{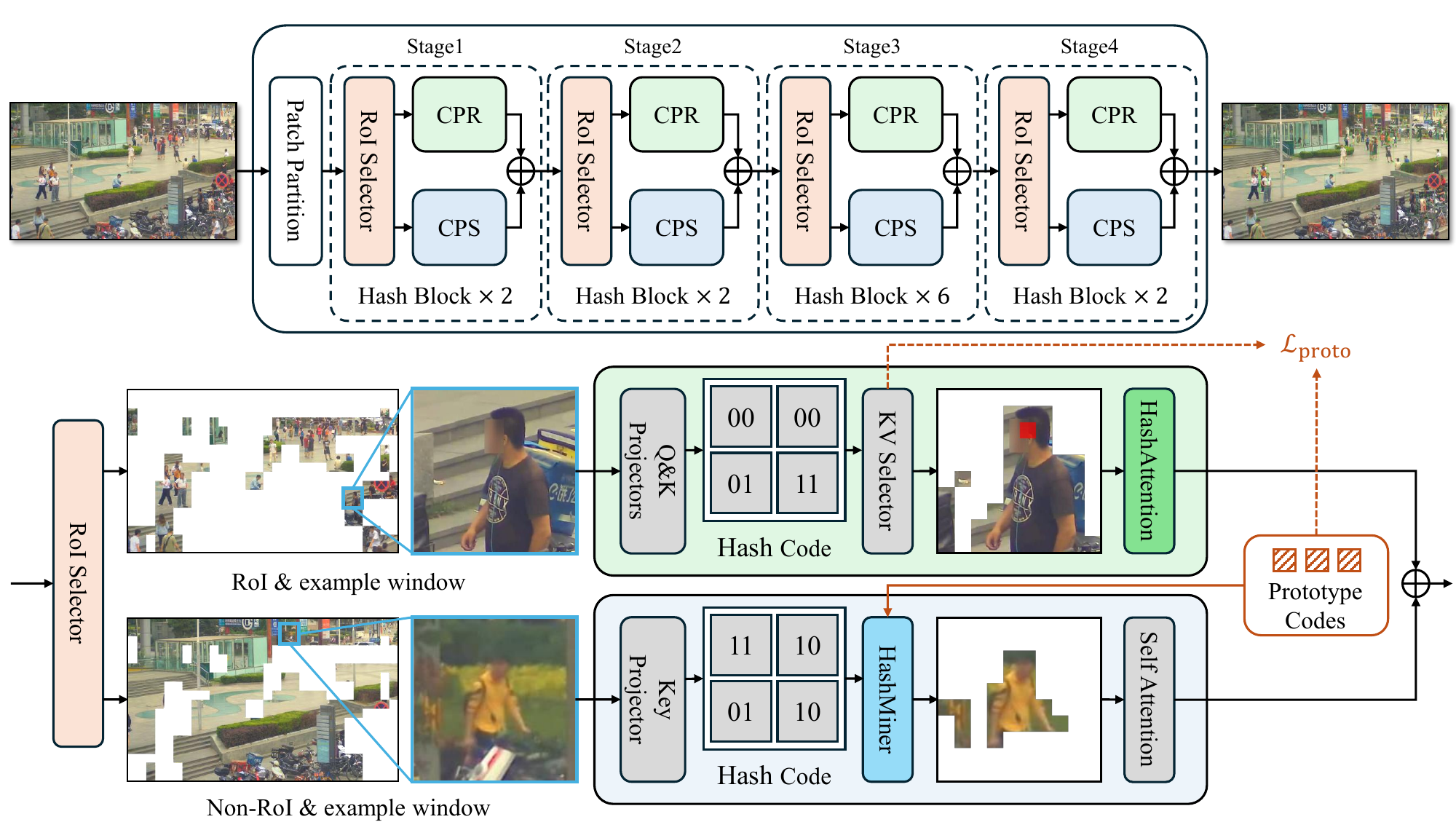}
    \caption{\textbf{Architecture overview of BiHashFormer.} CPR: \textbf{C}om\textbf{PR}ession branch. CPS: \textbf{C}om\textbf{P}en\textbf{S}ation branch. RoI Selector devides images into RoI for CPR and Non-RoI for CPS. Hash Code in the figure demonstrate an example of 2-bit for Eq.~\eqref{eq:9}. An example of HashAttention is listed in the green box: for the red query, pixels in the image are selected by KV Selector. Prototype codes are trained in CPR through $\mathcal{L}_{proto}$, then used in HashMiner. HashMiner selects tokens for further feature extraction in CPS.}
    \label{fig:2}
\end{figure*}

\section{Method}
\subsection{Overview of BiHashFormer}
In the main network, input image is partitioned into non-overlapping patches of size ($\frac{H}{4}\times \frac{W}{4} \times 48$) where $H$ and $W$ denote the original size of input. Through linear embedding layer, the number of channel is projected into $C$ in the first stage. In subsequent stages, the width and height of the feature maps are progressively halved while channel doubled by merging patches, reshaping the size into ($\frac{H}{8}\times \frac{W}{8} \times 2C$) for RoI Selector. The RoI Selector segments the feature map into $w\times w$ windows, further categorizing them into RoI and non-RoI regions based on variance, as shown in \Cref{fig:2}. RoIs are fed into the compression branch for feature extraction, which accounts for the major computational cost of the model; non-RoI regions are processed in the compensation branch, where features are first filtered and then extracted to recover information from small and hard targets. The features extracted from the two branches are directly merged. The RoI Selector together with the two branches forms a Hash Block, which serves as the basic unit of each stage. The model consists of four stages, with the number of Hash Blocks in each stage (i.e., the stage depth) set to $[2, 2, 6, 2]$.

In the compression branch, query and key will be separately projected into low dimensional Hamming space. HashAttention is then applied to RoIs window by window to extract features. Specifically, all queries participate in the attention computation, but the number of interacting key–value pairs is halved according to the Hamming distance. We illustrates the selected k-v pairs for an example query (colored red) in \Cref{fig:2}. Prototype codes is learned within this branch through $\mathcal{L}_{proto}$ (see Section~\ref{sec:prototypes}). In the compensation branch, only the keys are projected. HashMiner then identifies the target-associated tokens using prototype codes and applies conventional self-attention to extract features. An example window is shown in \Cref{fig:2}.

\subsection{RoI Selector}

\noindent\textbf{Variance-based token scoring.}
Identifying windows most likely to contain targets (i.e., RoIs) presents a critical challenge, which is the decisive factor for the computational efficiency of the compression branch. Observing that background regions (e.g., sky, plaza and roads) typically exhibit more uniform pixel distributions (i.e., lower variance) than object regions, we score each window using the variance of pixels. For the input feature map $\mathbf{F}$ in shape $(B\times H\times W\times C)$, we compute its average pooled result in shape $(B\times \frac{H}{w}\times \frac{W}{w}\times C)$ and interpolate it back first:
\begin{equation}
    \mathbf{F}_{\text{pooled}} =\text{AvgPool}_{w \times w}(\mathbf{F}), \quad \hat{\mathbf{F}} = \text{Interpolate}(\mathbf{F}_{\text{pooled}}, w)
    \label{eq:1}
\end{equation}

Here we obtain mean feature $\hat{\mathbf{F}}$, taking neighboring windows into account. Then we partition and flatten the residual for each window as below, where $k$ means $k^{th}$ window:
\begin{equation}
    \hat{\mathbf{R}} = \mathbf{F} - \hat{\mathbf{F}}, \quad \mathbf{W}_k = \text{Flatten}(\hat{\textbf{R}}_k)
    \label{eq:2}
\end{equation}

A simple MLP $f_{score}$ is employed to transform the token sequence into scores corresponding to each window, which are normalized subsequently:
\begin{equation}
    s_k = f_{score}(\mathbf{W}_k), \quad \hat{s}_k = \frac{\exp(s_k)}{\sum_{j=1}^{N_w} \exp(s_j)}
    \label{eq:3}
\end{equation}

\noindent\textbf{Top-k selection.}
Assigning too many windows to the RoI leads to a surge in the computational cost of the compression branch, whereas too few windows cause most targets to be routed to the compensation branch, resulting in insufficient feature extraction. To this end, we use $\hat{\rho}$ generated by a MLP to select the N highest-scoring windows for the CPR:
\begin{equation}
    N = \lceil \hat{\rho} \cdot N_w \rceil
    \label{eq:4}
\end{equation}
Here $N_w$ is the total number of winodws and $N$ highest-scoring windows will be assigned as RoI. Taking $N_w=110,\hat{\rho}=0.15$ as an example, $N=\lceil 0.15 \cdot 110 \rceil=17$, thus top-17 windows ranked by score will be sent to the compression branch and the rest for compensation.

\noindent\textbf{Generating $\hat{\rho}$.}
The foreground proportion $\rho$ measures spatial complexity. For boxes $\text{BBox}_i$ in the input image coordinate system and image region $\Omega$ of size $H\times W$, it is the clipped union area divided by image area:
\begin{equation}
    \rho = \frac{\operatorname{Area}\!\left(\bigcup_i (\text{BBox}_i\cap\Omega)\right)}{H \cdot W}
    \label{eq:5}
\end{equation}

In order to obtain such a feature of the overall image, we used feature map for extraction through a MLP. Because the MLP’s input dimensions are fixed and restricted to one-dimensional vectors, we apply a size-normalization operation to the feature maps:
\begin{equation}
    \mathbf{F}_{\text{pooled}} = \text{AdpPool}(\mathbf{F}), \quad \mathbf{F}_{\text{flat}} = \text{Flatten}(\mathbf{F}_{\text{pooled}})
    \label{eq:6}
\end{equation}
The adaptive average pooling (AdpPool) downsamples the input feature map $\mathbf{F}$ (derived from the backbone) to a predetermined, fixed spatial resolution regardless of the original size of input images. Here the size for 4 stages is $[32^2,\ 16^2,\ 8^2,\ 4^2]$. The subsequent Flatten layer then serializes the two-dimensional output $\mathbf{F}_{\text{pooled}}$ into a one-dimensional feature vector $\mathbf{F}_{\text{flat}}$. The dimensionality of this vector is inherently determined by the output size chosen for the adaptive pooling and the number of feature channels $C$, and is strategically configured based on the specific stage within the BiHashFormer backbone where RoI Selector is integrated (larger dimensions for early layers capturing fine details, smaller for deeper semantic layers).

After obtaining the $\mathbf{F}_{\text{flat}}$, we employ a lightweight two-layer MLP, $f_{\text{pred}}$, to predict the foreground proportion $\hat{\rho}$ for RoI Selector in each stage as shown in \Cref{fig:2}, marked as stage $t$:
\begin{equation}
    \hat{\rho^t} = \sigma(f_{\text{pred}}(\mathbf{F}^t_{\text{flat}}))
    \label{eq:7}
\end{equation}
where layer of Sigmoid ensures that $\hat{\rho}$ lies between 0 and 1. $\hat{\rho}$ will be used for top-k selection, shown in \Cref{eq:4}. The MLP is trained by $\mathcal{L}_{select}$ below, where $\gamma$ lays in the interval $(0,1)$, motivating smaller $\hat{\rho}$ through $L_1$ penalty. We set $\gamma$ to 0.1 through experiments.
\begin{equation}
    \mathcal{L}_{\text{select}} = (\rho - \hat{\rho})^2 + \gamma \cdot \hat{\rho}
    \label{eq:8}
\end{equation}

\subsection{HashAttention}

\noindent\textbf{Q\&K projectors.}
To evaluate token similarity efficiently in HRW detection, we project queries and keys into a low dimensional Hamming space. Compared with cosine similarity, its compact binary codes allow inexpensive integer operations:
\begin{equation}
    \begin{aligned}
        z = \mathbf{P}_{Q/K}^T\cdot x, \quad b = \mathbf{1}[z>0]
    \end{aligned}
    \label{eq:9}
\end{equation}
Here $x\in\mathrm{R}^d$ is a query or key, $\mathbf{P}_{Q/K}\in\mathrm{R}^{d\times k}$ and $k=32$. Separate Q/K projectors produce elementwise binary codes $b\in\{0,1\}^k$, with zero mapped to bit 0. Gradients use the Straight-Through Estimator (STE):
\begin{equation}
    \frac{\partial \mathcal{L}}{\partial z} =: \frac{1}{2} \frac{\partial \mathcal{L}}{\partial b}
    \label{eq:10}
\end{equation}
Notably, instead of projecting the raw tokens, we project the queries and keys. This is explained below.

\noindent\textbf{Compress KV for each Q.}
A larger window size can better accommodate complex HRW scenarios. However, for the tokens within each window, this implies greater interference when computing attention. We expect each token to engage in attention interactions only with the tokens it truly requires. This implies that we must know what information each token needs, as well as what information it contains. This precisely corresponds to the query and key in the attention mechanism, thereby explaining why we choose them for projection.
We employ the Hamming distance to measure the similarity between keys and queries:
\begin{equation}
    dis_H(b_q,b_k) = ||b_q \oplus b_k||_1
    \label{eq:11}
\end{equation}
Here $\oplus$ denotes bitwise XOR, and $||\cdot||_1$ represents the count of different positions. A larger Hamming distance indicates that the two vectors are farther apart in the Hamming space, implying a greater discrepancy between the information required by the query and the information contained in the key.
For a given query $q$, KV Selector selects the half of the keys with the smallest Hamming distance in the window, marked as $\mathbf{K}_{0.5}$. Each value is bound to the corresponding key from the same token, and compute the HashAttention (HA) as:
\begin{equation}
    \text{HA}(q, \mathbf{K}_{0.5}, \mathbf{V}_{0.5}) = \text{softmax}\left(\frac{q\mathbf{K}_{0.5}^{\top}}{\sqrt{d_{\text{key}}}}\right)\mathbf{V}_{0.5}
    \label{eq:12}
\end{equation}
HA is calculated in windows. Taking the case in \Cref{fig:2} as an example, we illustrate the projection results of four keys, simplified into a 2-bit form: 00, 00, 01, and 11. Considering a query projected as 11, it will select the key–value pairs corresponding to 01 and 11 (they get smaller $dis_H$ as 1 and 0), and compute HashAttention in the original space.

\subsection{HashMiner}

\noindent\textbf{K projector.}
To accommodate the characteristic of HRW scenarios where the foreground occupies only a small proportion, most windows are designated as non-RoI to reduce computational cost, as illustrated in \Cref{fig:2}. However, small or difficult targets are often submerged within non-RoI regions, calling for an efficient method to extract.

As mentioned earlier, since the key contains the intrinsic information of the token, it is well-suited for selection. In the compression branch, the K projector has already been sufficiently trained. Therefore, we directly adopt the pre-trained projection matrix $\mathbf{P}_K$ to obtain the low dimensional Hamming features of the keys in the compensation branch (follows \Cref{eq:9}), thereby achieving parameter reduction. In this branch, the parameters of $\mathbf{P}_K$ are frozen.

\noindent\textbf{Selection criteria.}
After obtaining the mapping of keys in the low dimensional Hamming space, we further need to identify which codes correspond to the semantics of “person”. After obtaining these cluster centers $c_i$, HashMiner can rapidly filter out the tokens requiring feature compensation based on the given radius $r=4$:
\begin{equation}
    \mathcal{X}_{comp} = \left\{x\;\middle|\;\min_{1\leq i\leq N_c} dis_H(b_K(x),c_i)\leq r\right\}
    \label{eq:13}
\end{equation}
Here $b_K(x)$ is the binary code of token $x$'s key and $N_c=3$ is the number of prototype codes, chosen by ablation. The following subsection describes their training.

\subsection{Prototype codes}
\label{sec:prototypes}

\noindent\textbf{Learning from the compression branch.}
Prototype codes are obtained by binarizing several trainable tokens into 0–1 representations, which need training data. Since the two branches within the same Hash Block share the key projector, the Hamming space corresponding to the keys remains consistent. As the compression branch processes RoIs that already contain rich target features, where prototype codes can be efficiently learned from. We employ the keys selected by the KV Selector to train the prototype codes, which reduces computation. Moreover, visualization experiments (\Cref{fig:3}) demonstrate that the selected keys contain richer target information.

\begin{samepage}
\noindent\textbf{Training for prototype codes.}
Inspired by clustering methods in machine learning, we expect the clusters corresponding to prototype codes to be as compact as possible within classes and as dispersed as possible across classes. The intra-class loss is defined as follows:
\begin{equation}
        d_j = \min_{1\leq i\leq N_c}\frac{dis_H(b_{k_j},c_i)}{k}, \quad \mathcal{L}_{intra} = \frac{1}{N_{\text{key}}}\sum_{j=1}^{N_{\text{key}}}d_j
    \label{eq:14}
\end{equation}
\end{samepage}
$b_{k_j}$ is the binary code of selected key $j$, $c_i$ the $i$th prototype code, and $N_{\text{key}}$ the number of selected keys. We use $k=32$ and the Hamming distance in~\Cref{eq:11}.

According to the triangle inequality, ensuring that the distance between prototypes is greater than $2r$ guarantees that their covered spaces do not overlap. Based on this, we introduce an inter-class penalty term to encourage prototypes to encompass richer semantics:

\begin{equation}
    \mathcal{L}_{cross} = \frac{1}{N_c(N_c-1)}\sum_{p\neq t}\max{\{0,2r+1-||c_p-c_t||_1\}}
    \label{eq:15}
\end{equation}

Here $r=4$ and $N_c=3$ is the same as \Cref{eq:13}. At initialization, each code has approximately $k/N_c$ ones in a disjoint bit block and zeros elsewhere, separating the prototypes. For example, with $k=3,N_c=3$, the initialized codes are 100, 010 and 001. Prototype latents are optimized through their binary codes using the STE in~\Cref{eq:10}; inference uses hard binary codes:

\begin{equation}
    \mathcal{L}_{proto} = \mathcal{L}_{intra} + \mathcal{L}_{cross}
    \label{eq:16}
\end{equation}

\section{Experiment}
\begin{table*}[tbp]
    \caption{\textbf{Computational cost and accuracy comparison of different models on PANDA.} GFLOPs are calculated exclusively for the backbone and neck when applicable. ``F'', ``B'' and ``O'' denote foreground, background and overall respectively. Baseline sources: \cite{cai2018cascade,he2016deep,yang2019clustered,li2020density,fan2022speed,chen2022towards,lin2017focal,zhao2024detrs,dai2021dynamic,Swin,SparseFormer,DINO}.}
    \label{tab:1}
    \centering
    \setlength{\tabcolsep}{2.5pt}
\begin{fullwidthtabular}{9}{llrrrrrrr}
    \toprule
         Method & Backbone & \multicolumn{3}{c}{GFLOPs} & \multicolumn{4}{c}{AP} \\
         \cmidrule(lr){3-5}\cmidrule(lr){6-9}
         & & F & B & O & 50 & small & medium & large \\
         \midrule
         CascadeRCNN  & ResNet-101  & 15.54 & 295.24 & 310.78 & - & 0.227 & 0.579 & 0.765\\
         ClusDet  & ResNet-50  & 10.35 & 196.71 & 207.07 & 0.718 & 0.219 & 0.696 & 0.782\\
         DMNet  & ResNet-50 & 10.35 & 196.71 & 207.07 & 0.540 & 0.119 & 0.371 & 0.714\\
         FasterRCNN  & ResNet-50 & 10.35 & 196.71 & 207.07 & 0.705 & 0.203 & 0.712 & 0.760\\
         GigaDet  & CSP-DarkNet-53 & \textbf{4.61} & 87.59 & 92.20 & 0.684 & 0.210 & 0.599 & 0.762\\
         PAN  & ResNet-50 & 10.35 & 196.71 & 207.07 & 0.715 & 0.256 & 0.719 & 0.768\\
         RetinaNet  & ResNet-101 & 15.77 & 299.62 & 315.39 & - & 0.221 & 0.561 & 0.740\\
         RT-DETR  & ResNet-50 & 10.35 & 196.71 & 207.07 & 0.612 & 0.137 & 0.532 & 0.735\\
         \midrule
         Dynamic-Head  & Swin-T  & \underline{5.74} & 109.1 & 114.8 & 0.592 & 0.165 & 0.537 & 0.694\\
         Dynamic-Head & SparseFormer  & 6.29 & 58.35 & 64.64 & 0.771 & 0.364 & 0.740 & \textbf{0.863}\\
         \rowcolor{RoyalBlue!20} Dynamic-Head & BiHashFormer (Ours) & 7.11 & \textbf{38.35} & \textbf{45.46} & \underline{0.799} & \underline{0.565} & \underline{0.783} & 0.841\\
         \midrule
         DINO  & ResNet-50  & 6.21 & 118.02 & 124.24 & 0.542 & 0.289 & 0.530 & 0.592\\
         DINO  & Swin-T  & 6.64 & 126.19 & 132.84 & 0.606 & 0.367 & 0.612 & 0.649\\
         DINO & SparseFormer  & 6.90 & 68.81 & 75.71 & 0.780 & 0.508 & 0.781 & 0.823\\
         \rowcolor{RoyalBlue!20} DINO & BiHashFormer (Ours) & 7.30 & \underline{40.01} & \underline{47.31} & \textbf{0.810} & \textbf{0.591} & \textbf{0.792} & \underline{0.847}\\
         \bottomrule
    \end{fullwidthtabular}
\end{table*}

\subsection{Datasets}

Our primary evaluation is conducted on the PANDA \cite{PANDA} dataset, the first human-centric gigapixel-level benchmark featuring 13 training and 5 testing scenes with approximately 16 million annotated bounding boxes. The average image resolution of this dataset, 25,000 $\times$ 16,000 pixels, presents unprecedented computational demands, establishing a rigorous testbed for efficiency assessment.

\subsection{Evaluation metrics}

Model efficiency is quantified using FLOPs and standard COCO metrics, with particular emphasis on $\text{AP}_{50}$, $\text{AP}_\text{small}$ ($<96\times 96$), $\text{AP}_\text{medium}$ ($96\times 96 \sim 288\times 288$) and $\text{AP}_\text{large}$ ($>288\times 288$). We further record the FLOPs on foreground (FLOPs-F) and background (FLOPs-B) to provide a more fine‑grained evaluation of model. To ensure fair comparison, all detectors process uniformly cropped 1280×800 windows from test images. Given the dynamic token selection mechanism in  BiHashFormer, its FLOPs are estimated by averaging measurements across the test set. All implementations utilized the MMDetection \cite{chen2019mmdetection} framework under standardized 36-epoch training protocols.

\subsection{Data analysis on PANDA}

As presented in \Cref{tab:1}, traditional detectors consistently exceed 200 GFLOPs for backbone computation on PANDA \cite{PANDA}, whereas modern Transformer-based approaches like Dynamic-head \cite{dai2021dynamic} and DINO \cite{DINO} reduces computational costs to sub-100 GFLOPs. SparseFormer achieves further reduction to 75.71 GFLOPs through fixed-ratio window selection while delivering significant AP gains.

BiHashFormer is combined with Dynamic-head \cite{dai2021dynamic} and DINO \cite{DINO} to evaluate its performance. The results indicate that BiHashFormer achieves superior detection performance with significantly lower computation. Compared with SparseFormer \cite{SparseFormer}, our backbone achieves a 37.5\% reduction in computational cost on DINO while improving $\text{AP}_{50}$ by 3.85\%. In comparison with Swin-T \cite{Swin}, the reduction in computation reaches 64.4\%. Notably, employing BiHashFormer on DINO yields a 16.34\% improvement in $\text{AP}_{\text{small}}$ over SparseFormer, and this gain further increases to 55.22\% on Dynamic-Head. It can be observed that the proportion of foreground computation in BiHashFormer has significantly increased, rising from 9.1\% in SparseFormer to 15.43\%. This indicates that our backbone achieves better efficiency.

In addition, we record the computational cost proportion of the two branches in BiHashFormer: 38.12 GFLOPs for the compression branch and 8.29 GFLOPs for the other. It can be observed that the compression branch constitutes the major computational overhead. Although the compensation branch receives a larger input region (85\% of all), it accounts for only 17.9\% of the branches computation, which highlights the efficiency of HashMiner’s selection.

\subsection{Study on HashAttention}

\begin{figure}
    \begin{minipage}[b]{0.49\linewidth}
    \centering
    \includegraphics[width=1\linewidth]{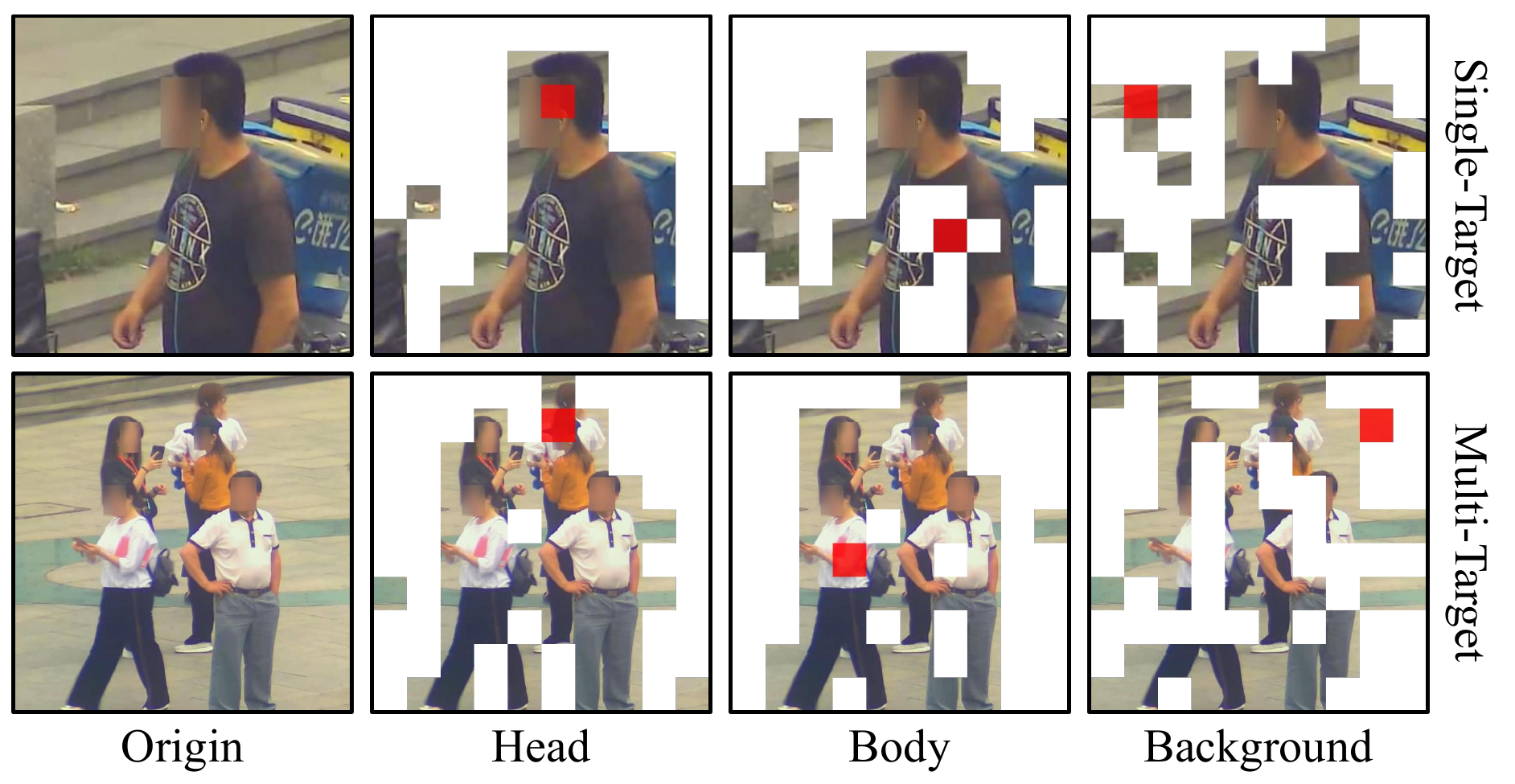}
    \caption{\textbf{Comparison of interacted KV for the chosen Q.} The pixel in red represent the query. Rows represent two target densities, while columns correspond to three types of queries.}
    \label{fig:3}
    \end{minipage}
    \hfill
    \begin{minipage}[b]{0.49\linewidth}
    \captionof{table}{\textbf{AP and FLOPs with different selection strategies.} FLOPs: calculation of the compression branch. AP denotes the performance of the complete model. $A\to \rho\%\ B$: using all $A$, selecting $\rho\%\ B$ for each $A$. HA: HashAttention. Att: classical attention. M/L: medium/large.}
    \label{tab:2}
    \centering
    \setlength{\tabcolsep}{2pt}
\begin{tabular*}{\linewidth}[b]{@{\extracolsep{\fill}}ccccc@{}}
        \toprule
        Selection & GFLOPs & $\mathrm{AP}_{50}$ & $\mathrm{AP}_{M}$ & $\mathrm{AP}_{L}$ \\
        \midrule
        $K\to 50\%\ Q$ & 38.14 & 0.785 & 0.775 & 0.815 \\
        \midrule
        $Q\to 20\%\ K$ & \textbf{16.97} & \underline{0.803} & 0.788 & 0.836 \\
        $Q\to 50\%\ K$ & \underline{38.12} & \textbf{0.810} & \underline{0.792} & \textbf{0.847} \\
        $Q\to 80\%\ K$ & 60.28 & 0.799 & \textbf{0.793} & 0.828 \\
        $Q\to 100\%\ K$ & 80.11 & 0.797 & 0.783 & \underline{0.837} \\
        \bottomrule
        \end{tabular*}
    \end{minipage}
\end{figure}

\textbf{Visualization of HashAttention.} We selected several scenarios with different target densities and extracted representative queries from each, visualizing their interacted key–value pairs as shown in \Cref{fig:3}.
It can be observed that both foreground and background queries pay greater attention to the key–value pairs corresponding to humans. This strongly supports the effectiveness of employing the KV Selector to choose keys during prototype code training.

Further, the figure reveals an interesting pattern: queries from the head tend to interact with key–value pairs covering the entire body, whereas queries from the body and background focus more on the key–value pairs located at the boundary between foreground and background. This may be because the head, as a key marker of a person, serves as a suitable anchor for feature extraction, linking other parts of the body to form the entirety of the detection target. In contrast, the body and the surrounding environment focus more on differences, assisting the backbone in distinguishing between target and background features. Thereby backbone can separate them in the Hamming space better.

\noindent\textbf{Different mechanisms of attention.} For HashAttention, we need to verify whether it outperforms the classical Attention in HRW scenarios. In addition, the idea of selecting half of the keys for a query also requires examination. To this end, we retain Hamming distance as the criterion and explore different selection strategies. The strategies are marked as $A\to \rho\%\ B$: using all $A$, selecting $\rho\%\ B$ for each $A$. The classical attention can thus be denoted as $Q\to 100\%\ K$.

According to \Cref{tab:2}, HashAttention outperforms classical Attention in both computational cost and detection performance. Line $K\to 50\%\ Q$ demonstrates that the supply–demand relationship between Q and k is not symmetric, swapping the order results in a significant performance degradation although the computational cost remains nearly unchanged. Experiments conducted with different values of $\rho$ indicate that 50\% is an appropriate choice.

\subsection{Study on the compensation branch}

\textbf{Necessity of the compensation branch.} Since the model requires at least one inference pathway and the compression branch serves as the main branch of BiHashFormer, we conduct an ablation study on the alternative compensation branch. First, we disabled this branch during inference. Although this reduced its computational cost, it significantly degraded detection performance, particularly for small objects, as evidenced by nearly a 51.4\% drop in $\text{AP}_{\text{small}}$ in Table X3 in the supplementary material.

Subsequently, we replaced this branch with a classical attention module for feature compensation. Specifically, the entire image was first down-sampled via average pooling with a reduction factor equal to the window size $w$. Attention computation was then performed on the down-sampled image, and the results from non-RoI regions were mapped back to the original resolution, forming a complete feature map together with the features obtained from the compression branch. This ensures that the compression branch remains unaffected while keeping the computational cost of the replacement comparable to the compensation branch. We define the efficiency as following to demonstrate the efficiency of different methods in compensating features for small objects, where $\text{AP}_{\text{small}}^{\text{NA}}$ refers to the 0.287 in Table X3:
\begin{equation}
    \text{Eff}=100\cdot (\text{AP}_{\text{small}}-\text{AP}_{\text{small}}^{\text{NA}})/\text{GFLOPs}
    \label{eq:17}
\end{equation}

These two ablation experiments demonstrate the necessity of the compensation branch: it cannot be replaced by simple feature extraction methods, nor can it be directly removed.

\noindent\textbf{Selection methods compared with HashMiner.} We aim to determine whether the effect brought by HashMiner is truly efficient or merely redundant. To this end, we experimented with several alternative selection methods and focused on comparing their computational cost, latency and $\text{AP}_{\text{small}}$.

The results in Table X1 show that whether using random selection, training a simple MLP to decide selection, or leveraging gradient-based criteria, none of these approaches outperform HashMiner in compensating features for small objects, proved by $\text{AP}_{\text{small}}$. Moreover, both computational cost and latency remaining well-controlled on HashMiner.

\subsection{Ablation study}

\textbf{Ablation study on adaptive sparsity rates.} SparseFormer \cite{SparseFormer} also adopts the concept of sparsity rate, but it uses a fixed parameter preset to 0.7. We choose this parameter from $[0.05,0.147,0.3,0.5,0.7,0.9]$ for comparison, where $0.147$ is the average $\hat{\rho}$ on BiHashFormer. The experiments are conducted with both DINO \cite{DINO} and Dynamic-Head \cite{dai2021dynamic}.

\begin{wrapfigure}{R}{0.5\linewidth}
    \centering
    \includegraphics[width=1\linewidth]{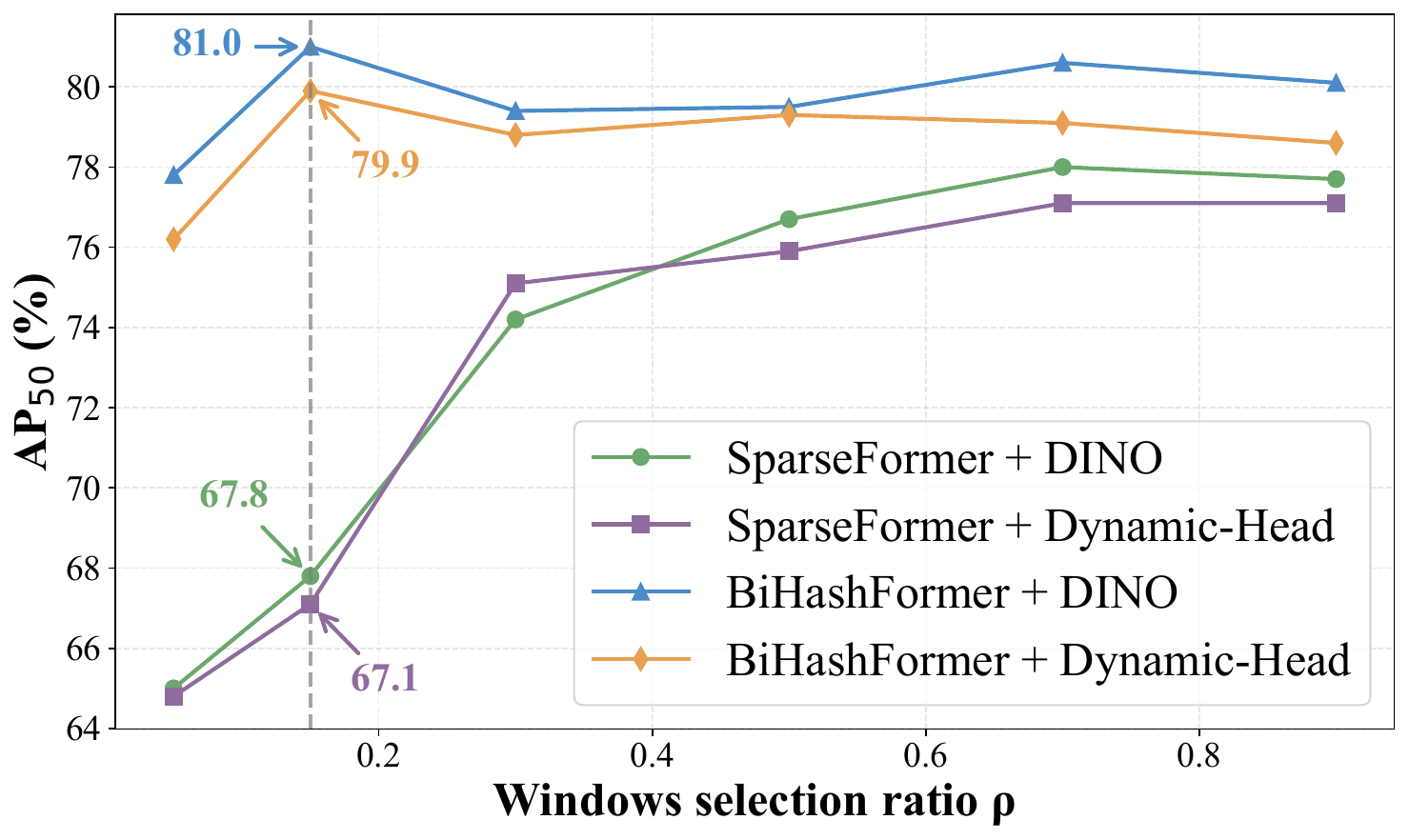}
    \caption{\textbf{AP$_{50}$ vs. windows selection ratio $\rho$ on SparseFormer and BiHashFormer.} The values annotated correspond to the $\text{AP}_{50}$ at $\rho = 0.147$.}
    \label{fig:4}
\end{wrapfigure}

As shown in \Cref{fig:4}, thanks to the compensation branch, our model performs well even at small values of $\rho$, whereas the detection accuracy of SparseFormer drops significantly. With the same sparsity rate as BiHashFormer, its AP$_{50}$ is consistently lower. This highlights the importance of dynamic sparsity in addressing target number variations in HRW scenarios.

\noindent\textbf{Ablation study on the dimension $k$.} The dimension $k$ of the projected Hamming space influences the two branches. It determines how much semantic information is contained in each bit of the hash code. An excessively large $k$ leads to overly fine semantic distinctions, undermining the purpose of a low-rank feature space. An excessively small $k$ results in semantic entanglement, making it difficult to identify clear boundaries. This is validated in \Cref{tab:3}.

\noindent\textbf{Ablation study on $\gamma$.} We select five values of $\gamma$ for ablation. Increasing $\gamma$ reduces computational cost at the expense of AP50, highlighting the trade-off between efficiency and accuracy. This is shown in Table X4 in the supplementary material.

\begin{table}
    \begin{minipage}[t]{0.49\linewidth}
    \caption{\textbf{$\text{AP}$ and FLOPs with different $k$.} Smaller $k$ leads to lower efficiency and worse results.}
    \label{tab:3}
    \centering
    \setlength{\tabcolsep}{2pt}
\begin{tabular*}{\linewidth}{@{\extracolsep{\fill}}ccccc@{}}
        \toprule
        $k$ & F & B & O & $\mathrm{AP}_{50}$ \\
        \midrule
        16 & \textbf{6.84} & 48.17 & 55.01 & 0.801 \\
        32 & 7.30 & 40.01 & 47.31 & \textbf{0.810} \\
        64 & 7.00 & \textbf{39.95} & \textbf{46.95} & 0.805 \\
        \bottomrule
        \end{tabular*}
    \end{minipage}
    \hfill
    \begin{minipage}[t]{0.49\linewidth}
    \caption{\textbf{$\text{AP}$ and FLOPs with different prototype numbers.} GFLOPs-CPS: compensation branch only.}
    \label{tab:4}
    \centering
    \setlength{\tabcolsep}{2pt}
\begin{tabular*}{\linewidth}{@{\extracolsep{\fill}}ccccc@{}}
        \toprule
        $N_c$ & F & CPS & $\mathrm{AP}_{50}$ & $\mathrm{AP}_{\mathrm{small}}$ \\
        \midrule
        1 & \textbf{7.12} & \textbf{5.86} & 0.807 & 0.498 \\
        3 & 7.30 & 40.01 & \textbf{0.810} & \textbf{0.591} \\
        5 & 7.29 & 17.41 & 0.806 & 0.582 \\
        \bottomrule
        \end{tabular*}
    \end{minipage}
\end{table}

\noindent\textbf{Ablation study on the window size $w$.} To handle the more complex scenarios and larger scale variations in HRW shots, we increase the window size, thereby allowing attention interactions among tokens to cover a broader range. In Swin, increasing $w$ leads to a sharp rise in the computation of attention. We alleviate this issue through RoI Selector and HashAttention. Among the four tested values $[7,\ 14,\ 21,\ 28]$, the best-performing setting is 21, as reported in Table X2.

\noindent\textbf{Ablation study on the number of prototype codes.} Similar to the K-Means clustering method, the number of prototype codes has a substantial impact on the final clustering performance. According to the results in \Cref{tab:4}, we set this parameter to 3 for the optimal effect of the compensation branch. The prototypes retrieve foreground tokens. The detector head predicts their categories.

\subsection{Supplementary experiments}

Section 2 of the supplementary material reports inference latency, DOTA/VisDrone generalization, and STE training stability. On COCO, BiHashFormer with Mask R-CNN and a $3\times$ schedule reaches 48.3 AP using three prototypes, versus 46.0 for Swin-T, with 35.6\% fewer backbone/neck FLOPs. Across 80 categories, three prototypes give the best accuracy among the tested counts, supporting class-agnostic foreground retrieval. With the same DINO head, BiHashFormer reaches 0.810 $\text{AP}_{50}$ at 47.31 GFLOPs and 2.8\,s, outperforming the complete Linformer and XAttention configurations. With matched DINOv3~\cite{simeoni2025dinov3} ViT-S+/16 initialization and 36-epoch training, it gains 4.0 $\text{AP}_{50}$ and 11.6 $\text{AP}_{\text{small}}$ points over the baseline with 64.4\% fewer FLOPs. On eight RTX 4090 GPUs with one image per GPU, training takes 19.7\,h versus SparseFormer's 23.2\,h and reaches 0.75 $\text{AP}_{50}$ five epochs earlier. Full protocols and results appear in supplementary Tables X9--X13.

\section{Conclusion}
By deeply integrating hashing concepts with object detection, we propose BiHashFormer, a dual-branch vision transformer. It employs RoI Selector to sparsely select RoI, leverages Hamming distance to refine the attention mechanism into a sparsified HashAttention and utilizes HashMiner to dynamically compensate for object features in non-RoI regions. Thereby it reduces FLOPs by 64.4\% versus Swin-T \cite{Swin} and 37.5\% versus SparseFormer \cite{SparseFormer} while achieving 81.0\% $\text{AP}_{50}$ on PANDA \cite{PANDA}. This dual-branch paradigm establishes a new standard for efficient large-scale vision systems, particularly for real-time applications with constrained resources.

\clearpage
\begingroup
\begin{ack}
This work was supported in part by the National Natural Science Foundation of China (NSFC; Nos. 82441013 and 62572268);
the National Science and Technology Major Project (No. 2022ZD0119402);
the Ministry of Science and Technology of China (contract No. 2024YFB2809103);
the Fundamental and Interdisciplinary Disciplines Breakthrough Plan of the Ministry of Education of China (No. JYB2025XDXM904);
the ``Pioneer'' and ``Leading Goose'' R\&D Program of Zhejiang (No. 2024C01142);
and the Tsinghua University -- Migu Xinkong Culture Technology (Xiamen) Co., Ltd. Joint Research Center for Intelligent Light Field and Interaction Technology, Tsinghua University - Joint R\&D Project (R2511DKW) -- Phase 2.
We also acknowledge the KLATASDS-MOE.
\end{ack}
\endgroup

\bibliographystyle{named}
\bibliography{neurips_2026}


\clearpage
\section*{NeurIPS Paper Checklist}

\begin{enumerate}

\item {\bf Claims}
    \item[] Question: Do the main claims made in the abstract and introduction accurately reflect the paper's contributions and scope?
    \item[] Answer: \answerYes{} 
    \item[] Justification: The main claims can accurately reflect the paper's contributions and scope.
    \item[] Guidelines:
    \begin{itemize}
        \item The answer \answerNA{} means that the abstract and introduction do not include the claims made in the paper.
        \item The abstract and/or introduction should clearly state the claims made, including the contributions made in the paper and important assumptions and limitations. A \answerNo{} or \answerNA{} answer to this question will not be perceived well by the reviewers. 
        \item The claims made should match theoretical and experimental results, and reflect how much the results can be expected to generalize to other settings. 
        \item It is fine to include aspirational goals as motivation as long as it is clear that these goals are not attained by the paper. 
    \end{itemize}

\item {\bf Limitations}
    \item[] Question: Does the paper discuss the limitations of the work performed by the authors?
    \item[] Answer: \answerYes{} 
    \item[] Justification: The reason behind a pattern in section 4.4 need further investigation.
    \item[] Guidelines:
    \begin{itemize}
        \item The answer \answerNA{} means that the paper has no limitation while the answer \answerNo{} means that the paper has limitations, but those are not discussed in the paper. 
        \item The authors are encouraged to create a separate ``Limitations'' section in their paper.
        \item The paper should point out any strong assumptions and how robust the results are to violations of these assumptions (e.g., independence assumptions, noiseless settings, model well-specification, asymptotic approximations only holding locally). The authors should reflect on how these assumptions might be violated in practice and what the implications would be.
        \item The authors should reflect on the scope of the claims made, e.g., if the approach was only tested on a few datasets or with a few runs. In general, empirical results often depend on implicit assumptions, which should be articulated.
        \item The authors should reflect on the factors that influence the performance of the approach. For example, a facial recognition algorithm may perform poorly when image resolution is low or images are taken in low lighting. Or a speech-to-text system might not be used reliably to provide closed captions for online lectures because it fails to handle technical jargon.
        \item The authors should discuss the computational efficiency of the proposed algorithms and how they scale with dataset size.
        \item If applicable, the authors should discuss possible limitations of their approach to address problems of privacy and fairness.
        \item While the authors might fear that complete honesty about limitations might be used by reviewers as grounds for rejection, a worse outcome might be that reviewers discover limitations that aren't acknowledged in the paper. The authors should use their best judgment and recognize that individual actions in favor of transparency play an important role in developing norms that preserve the integrity of the community. Reviewers will be specifically instructed to not penalize honesty concerning limitations.
    \end{itemize}

\item {\bf Theory assumptions and proofs}
    \item[] Question: For each theoretical result, does the paper provide the full set of assumptions and a complete (and correct) proof?
    \item[] Answer: \answerYes{} 
    \item[] Justification: The proof is correct.
    \item[] Guidelines:
    \begin{itemize}
        \item The answer \answerNA{} means that the paper does not include theoretical results. 
        \item All the theorems, formulas, and proofs in the paper should be numbered and cross-referenced.
        \item All assumptions should be clearly stated or referenced in the statement of any theorems.
        \item The proofs can either appear in the main paper or the supplemental material, but if they appear in the supplemental material, the authors are encouraged to provide a short proof sketch to provide intuition. 
        \item Inversely, any informal proof provided in the core of the paper should be complemented by formal proofs provided in appendix or supplemental material.
        \item Theorems and Lemmas that the proof relies upon should be properly referenced. 
    \end{itemize}

    \item {\bf Experimental result reproducibility}
    \item[] Question: Does the paper fully disclose all the information needed to reproduce the main experimental results of the paper to the extent that it affects the main claims and/or conclusions of the paper (regardless of whether the code and data are provided or not)?
    \item[] Answer: \answerYes{} 
    \item[] Justification: Hyper-parameters are given.
    \item[] Guidelines:
    \begin{itemize}
        \item The answer \answerNA{} means that the paper does not include experiments.
        \item If the paper includes experiments, a \answerNo{} answer to this question will not be perceived well by the reviewers: Making the paper reproducible is important, regardless of whether the code and data are provided or not.
        \item If the contribution is a dataset and\slash or model, the authors should describe the steps taken to make their results reproducible or verifiable. 
        \item Depending on the contribution, reproducibility can be accomplished in various ways. For example, if the contribution is a novel architecture, describing the architecture fully might suffice, or if the contribution is a specific model and empirical evaluation, it may be necessary to either make it possible for others to replicate the model with the same dataset, or provide access to the model. In general. releasing code and data is often one good way to accomplish this, but reproducibility can also be provided via detailed instructions for how to replicate the results, access to a hosted model (e.g., in the case of a large language model), releasing of a model checkpoint, or other means that are appropriate to the research performed.
        \item While NeurIPS does not require releasing code, the conference does require all submissions to provide some reasonable avenue for reproducibility, which may depend on the nature of the contribution. For example
        \begin{enumerate}
            \item If the contribution is primarily a new algorithm, the paper should make it clear how to reproduce that algorithm.
            \item If the contribution is primarily a new model architecture, the paper should describe the architecture clearly and fully.
            \item If the contribution is a new model (e.g., a large language model), then there should either be a way to access this model for reproducing the results or a way to reproduce the model (e.g., with an open-source dataset or instructions for how to construct the dataset).
            \item We recognize that reproducibility may be tricky in some cases, in which case authors are welcome to describe the particular way they provide for reproducibility. In the case of closed-source models, it may be that access to the model is limited in some way (e.g., to registered users), but it should be possible for other researchers to have some path to reproducing or verifying the results.
        \end{enumerate}
    \end{itemize}

\item {\bf Open access to data and code}
    \item[] Question: Does the paper provide open access to the data and code, with sufficient instructions to faithfully reproduce the main experimental results, as described in supplemental material?
    \item[] Answer: \answerNo{} 
    \item[] Justification: Codes in PyTorch are given in the supplementary material. But the complete code has not been released yet.
    \item[] Guidelines:
    \begin{itemize}
        \item The answer \answerNA{} means that paper does not include experiments requiring code.
        \item Please see the NeurIPS code and data submission guidelines (\url{https://neurips.cc/public/guides/CodeSubmissionPolicy}) for more details.
        \item While we encourage the release of code and data, we understand that this might not be possible, so \answerNo{} is an acceptable answer. Papers cannot be rejected simply for not including code, unless this is central to the contribution (e.g., for a new open-source benchmark).
        \item The instructions should contain the exact command and environment needed to run to reproduce the results. See the NeurIPS code and data submission guidelines (\url{https://neurips.cc/public/guides/CodeSubmissionPolicy}) for more details.
        \item The authors should provide instructions on data access and preparation, including how to access the raw data, preprocessed data, intermediate data, and generated data, etc.
        \item The authors should provide scripts to reproduce all experimental results for the new proposed method and baselines. If only a subset of experiments are reproducible, they should state which ones are omitted from the script and why.
        \item At submission time, to preserve anonymity, the authors should release anonymized versions (if applicable).
        \item Providing as much information as possible in supplemental material (appended to the paper) is recommended, but including URLs to data and code is permitted.
    \end{itemize}

\item {\bf Experimental setting/details}
    \item[] Question: Does the paper specify all the training and test details (e.g., data splits, hyperparameters, how they were chosen, type of optimizer) necessary to understand the results?
    \item[] Answer: \answerYes{} 
    \item[] Justification: They are in the section 4.
    \item[] Guidelines:
    \begin{itemize}
        \item The answer \answerNA{} means that the paper does not include experiments.
        \item The experimental setting should be presented in the core of the paper to a level of detail that is necessary to appreciate the results and make sense of them.
        \item The full details can be provided either with the code, in appendix, or as supplemental material.
    \end{itemize}

\item {\bf Experiment statistical significance}
    \item[] Question: Does the paper report error bars suitably and correctly defined or other appropriate information about the statistical significance of the experiments?
    \item[] Answer: \answerYes{} 
    \item[] Justification: Some experiments take the average value of many runs.
    \item[] Guidelines:
    \begin{itemize}
        \item The answer \answerNA{} means that the paper does not include experiments.
        \item The authors should answer \answerYes{} if the results are accompanied by error bars, confidence intervals, or statistical significance tests, at least for the experiments that support the main claims of the paper.
        \item The factors of variability that the error bars are capturing should be clearly stated (for example, train/test split, initialization, random drawing of some parameter, or overall run with given experimental conditions).
        \item The method for calculating the error bars should be explained (closed form formula, call to a library function, bootstrap, etc.)
        \item The assumptions made should be given (e.g., Normally distributed errors).
        \item It should be clear whether the error bar is the standard deviation or the standard error of the mean.
        \item It is OK to report 1-sigma error bars, but one should state it. The authors should preferably report a 2-sigma error bar than state that they have a 96\% CI, if the hypothesis of Normality of errors is not verified.
        \item For asymmetric distributions, the authors should be careful not to show in tables or figures symmetric error bars that would yield results that are out of range (e.g., negative error rates).
        \item If error bars are reported in tables or plots, the authors should explain in the text how they were calculated and reference the corresponding figures or tables in the text.
    \end{itemize}

\item {\bf Experiments compute resources}
    \item[] Question: For each experiment, does the paper provide sufficient information on the computer resources (type of compute workers, memory, time of execution) needed to reproduce the experiments?
    \item[] Answer: \answerYes{} 
    \item[] Justification: AGX Orin and RTX 4090 are used.
    \item[] Guidelines:
    \begin{itemize}
        \item The answer \answerNA{} means that the paper does not include experiments.
        \item The paper should indicate the type of compute workers CPU or GPU, internal cluster, or cloud provider, including relevant memory and storage.
        \item The paper should provide the amount of compute required for each of the individual experimental runs as well as estimate the total compute. 
        \item The paper should disclose whether the full research project required more compute than the experiments reported in the paper (e.g., preliminary or failed experiments that didn't make it into the paper). 
    \end{itemize}
    
\item {\bf Code of ethics}
    \item[] Question: Does the research conducted in the paper conform, in every respect, with the NeurIPS Code of Ethics \url{https://neurips.cc/public/EthicsGuidelines}?
    \item[] Answer: \answerYes{} 
    \item[] Justification: We reviewed the guidelines and found no conflicts.
    \item[] Guidelines:
    \begin{itemize}
        \item The answer \answerNA{} means that the authors have not reviewed the NeurIPS Code of Ethics.
        \item If the authors answer \answerNo, they should explain the special circumstances that require a deviation from the Code of Ethics.
        \item The authors should make sure to preserve anonymity (e.g., if there is a special consideration due to laws or regulations in their jurisdiction).
    \end{itemize}

\item {\bf Broader impacts}
    \item[] Question: Does the paper discuss both potential positive societal impacts and negative societal impacts of the work performed?
    \item[] Answer: \answerYes{} 
    \item[] Justification: They are both discussed.
    \item[] Guidelines:
    \begin{itemize}
        \item The answer \answerNA{} means that there is no societal impact of the work performed.
        \item If the authors answer \answerNA{} or \answerNo, they should explain why their work has no societal impact or why the paper does not address societal impact.
        \item Examples of negative societal impacts include potential malicious or unintended uses (e.g., disinformation, generating fake profiles, surveillance), fairness considerations (e.g., deployment of technologies that could make decisions that unfairly impact specific groups), privacy considerations, and security considerations.
        \item The conference expects that many papers will be foundational research and not tied to particular applications, let alone deployments. However, if there is a direct path to any negative applications, the authors should point it out. For example, it is legitimate to point out that an improvement in the quality of generative models could be used to generate Deepfakes for disinformation. On the other hand, it is not needed to point out that a generic algorithm for optimizing neural networks could enable people to train models that generate Deepfakes faster.
        \item The authors should consider possible harms that could arise when the technology is being used as intended and functioning correctly, harms that could arise when the technology is being used as intended but gives incorrect results, and harms following from (intentional or unintentional) misuse of the technology.
        \item If there are negative societal impacts, the authors could also discuss possible mitigation strategies (e.g., gated release of models, providing defenses in addition to attacks, mechanisms for monitoring misuse, mechanisms to monitor how a system learns from feedback over time, improving the efficiency and accessibility of ML).
    \end{itemize}
    
\item {\bf Safeguards}
    \item[] Question: Does the paper describe safeguards that have been put in place for responsible release of data or models that have a high risk for misuse (e.g., pre-trained language models, image generators, or scraped datasets)?
    \item[] Answer: \answerNA{} 
    \item[] Justification: N/A
    \item[] Guidelines:
    \begin{itemize}
        \item The answer \answerNA{} means that the paper poses no such risks.
        \item Released models that have a high risk for misuse or dual-use should be released with necessary safeguards to allow for controlled use of the model, for example by requiring that users adhere to usage guidelines or restrictions to access the model or implementing safety filters. 
        \item Datasets that have been scraped from the Internet could pose safety risks. The authors should describe how they avoided releasing unsafe images.
        \item We recognize that providing effective safeguards is challenging, and many papers do not require this, but we encourage authors to take this into account and make a best faith effort.
    \end{itemize}

\item {\bf Licenses for existing assets}
    \item[] Question: Are the creators or original owners of assets (e.g., code, data, models), used in the paper, properly credited and are the license and terms of use explicitly mentioned and properly respected?
    \item[] Answer: \answerYes{} 
    \item[] Justification: All reused assets are cited with their sources.
    \item[] Guidelines:
    \begin{itemize}
        \item The answer \answerNA{} means that the paper does not use existing assets.
        \item The authors should cite the original paper that produced the code package or dataset.
        \item The authors should state which version of the asset is used and, if possible, include a URL.
        \item The name of the license (e.g., CC-BY 4.0) should be included for each asset.
        \item For scraped data from a particular source (e.g., website), the copyright and terms of service of that source should be provided.
        \item If assets are released, the license, copyright information, and terms of use in the package should be provided. For popular datasets, \url{paperswithcode.com/datasets} has curated licenses for some datasets. Their licensing guide can help determine the license of a dataset.
        \item For existing datasets that are re-packaged, both the original license and the license of the derived asset (if it has changed) should be provided.
        \item If this information is not available online, the authors are encouraged to reach out to the asset's creators.
    \end{itemize}

\item {\bf New assets}
    \item[] Question: Are new assets introduced in the paper well documented and is the documentation provided alongside the assets?
    \item[] Answer: \answerNA{} 
    \item[] Justification: New assets are not released yet.
    \item[] Guidelines:
    \begin{itemize}
        \item The answer \answerNA{} means that the paper does not release new assets.
        \item Researchers should communicate the details of the dataset\slash code\slash model as part of their submissions via structured templates. This includes details about training, license, limitations, etc. 
        \item The paper should discuss whether and how consent was obtained from people whose asset is used.
        \item At submission time, remember to anonymize your assets (if applicable). You can either create an anonymized URL or include an anonymized zip file.
    \end{itemize}

\item {\bf Crowdsourcing and research with human subjects}
    \item[] Question: For crowdsourcing experiments and research with human subjects, does the paper include the full text of instructions given to participants and screenshots, if applicable, as well as details about compensation (if any)? 
    \item[] Answer: \answerNA{} 
    \item[] Justification: N/A
    \item[] Guidelines:
    \begin{itemize}
        \item The answer \answerNA{} means that the paper does not involve crowdsourcing nor research with human subjects.
        \item Including this information in the supplemental material is fine, but if the main contribution of the paper involves human subjects, then as much detail as possible should be included in the main paper. 
        \item According to the NeurIPS Code of Ethics, workers involved in data collection, curation, or other labor should be paid at least the minimum wage in the country of the data collector. 
    \end{itemize}

\item {\bf Institutional review board (IRB) approvals or equivalent for research with human subjects}
    \item[] Question: Does the paper describe potential risks incurred by study participants, whether such risks were disclosed to the subjects, and whether Institutional Review Board (IRB) approvals (or an equivalent approval/review based on the requirements of your country or institution) were obtained?
    \item[] Answer: \answerNA{} 
    \item[] Justification: N/A
    \item[] Guidelines:
    \begin{itemize}
        \item The answer \answerNA{} means that the paper does not involve crowdsourcing nor research with human subjects.
        \item Depending on the country in which research is conducted, IRB approval (or equivalent) may be required for any human subjects research. If you obtained IRB approval, you should clearly state this in the paper. 
        \item We recognize that the procedures for this may vary significantly between institutions and locations, and we expect authors to adhere to the NeurIPS Code of Ethics and the guidelines for their institution. 
        \item For initial submissions, do not include any information that would break anonymity (if applicable), such as the institution conducting the review.
    \end{itemize}

\item {\bf Declaration of LLM usage}
    \item[] Question: Does the paper describe the usage of LLMs if it is an important, original, or non-standard component of the core methods in this research? Note that if the LLM is used only for writing, editing, or formatting purposes and does \emph{not} impact the core methodology, scientific rigor, or originality of the research, declaration is not required.
    \item[] Answer: \answerNA{} 
    \item[] Justification: N/A
    \item[] Guidelines:
    \begin{itemize}
        \item The answer \answerNA{} means that the core method development in this research does not involve LLMs as any important, original, or non-standard components.
        \item Please refer to our LLM policy in the NeurIPS handbook for what should or should not be described.
    \end{itemize}

\end{enumerate}

\end{document}


\setcounter{footnote}{0}
\maketitle
\setcounter{footnote}{0}

\section{Supplementary tables}
\label{sec:supp-tables}

This section presents tables that are cited in the main text. \Cref{tab:X1} is cited in Section 4.5 (“Selection methods compared with HashMiner”). \Cref{tab:X2} is cited in Section 4.6 (“Ablation study on the window size $w$”). \Cref{tab:X3} is cited in Section 4.5 (“Necessity of the compensation branch”). \Cref{tab:X4} is cited in Section 4.6 (“Ablation study on $\gamma$”). We report all numbers from evaluation on PANDA.

\begin{table}[htbp]
    \caption{\textbf{$\text{AP}_{\text{small}}$, FLOPs and latency with different compensation methods.} FLOPs: selection cost only. Latency is measured as the average over four stages, 50 forward passes of the backbone..}
    \label{tab:X1}
    \centering
\begin{fullwidthtabular}{4}{cccc}
        \toprule
        Selection Method & MFLOPs & $\text{AP}_{\text{small}}$ & Latency(ms) \\
        \midrule
        Random & \textbf{0} & 0.291 & \textbf{0.134} \\
        MLP & 1803 & \underline{0.375} & 1.793 \\
        Gradient & \underline{251} & 0.336 & \underline{0.847} \\
        \rowcolor{RoyalBlue!20} HashMiner (Ours) & 1179 & \textbf{0.591} & 1.196 \\
        \bottomrule
        \end{fullwidthtabular}
\end{table}

\begin{table}[htbp]
    \caption{\textbf{$\text{AP}_{50}$, FLOPs and FPS with different $w$.} GFLOPs-CPR: compression branch only. FPS is evaluated on single RTX 4090.}
    \label{tab:X2}
    \centering
\begin{tabular*}{\linewidth}{@{\extracolsep{\fill}}ccccc@{}}
        \toprule
        Window Size $w$ & GFLOPs-F & GFLOPs-CPR & $\text{AP}_{50}$ & FPS \\
        \midrule
        7 & \textbf{6.11} & \textbf{8.92} & 0.792 & \textbf{1.425} \\
        14 & \underline{6.89} & \underline{22.21} & \underline{0.806} & \underline{0.811} \\
        21 & 7.30 & 38.12 & \textbf{0.810} & 0.356 \\
        28 & 7.17 & 73.86 & 0.803 & 0.174 \\
        \bottomrule
        \end{tabular*}
\end{table}

\begin{table}[htbp]
    \caption{\textbf{AP, FLOPs and Eff with different compensation methods.} FLOPs: compensation cost only. AP denotes the performance of the complete model and Eff is defined in Equation (17).}
    \label{tab:X3}
    \centering
\begin{fullwidthtabular}{5}{ccccc}
        \toprule
        Compensation Method & GFLOPs & $\text{AP}_{50}$ & $\text{AP}_{\text{small}}$ & Eff \\
        \midrule
        - & \textbf{0} & 0.786 & 0.287 & - \\
        Attention Module & 13.28 & 0.793 & 0.311 & 0.18 \\
        \rowcolor{RoyalBlue!20} CPS Branch (Ours) & 8.29 & \textbf{0.810} & \textbf{0.591} & \textbf{3.67} \\
        \bottomrule
        \end{fullwidthtabular}
\end{table}

\begin{table}[htbp]
    \caption{\textbf{AP$_{50}$, FLOPs and FPS across different $\gamma$.}}
    \label{tab:X4}
    \centering
\begin{tabular*}{\linewidth}{@{\extracolsep{\fill}}ccccc@{}}
        \toprule
        $\gamma$ & AP$_{50}$$\uparrow$ & GFLOPs-O$\downarrow$ & GFLOPs-F$\downarrow$ & FPS$\uparrow$ \\
        \midrule
        0.1 & 0.810 & 47.31 & 7.30 & 0.356 \\
        \midrule
        0.3 & 0.808 & 44.25 & 6.91 & 0.381 \\
        \midrule
        0.5 & 0.807 & 42.57 & 6.72 & 0.397 \\
        \midrule
        0.7 & 0.804 & 40.86 & 6.19 & 0.414 \\
        \midrule
        0.9 & 0.801 & 37.59 & 5.68 & 0.452 \\
        \bottomrule
        \end{tabular*}
\end{table}

\FloatBarrier
\section{Supplementary experiments (Section 4.7)}
\label{sec:supp-experiments}

\subsection{Inference latency}
We evaluated the inference time of different models on PANDA using NVIDIA AGX Orin and RTX 4090. The former is widely used on edge devices. As shown in \Cref{tab:X5} and \Cref{tab:X6}, our method is faster than SparseFormer \cite{SparseFormer} with higher AP$_{50}$. Compared with PAN \cite{fan2022speed}, we \textbf{improved AP$_{50}$ by 12\% and speed up by 4$\times$}, demonstrating its \textbf{efficiency and accuracy}. Per-component wall-clock latency for the AGX Orin: backbone 4.4s, DINO head 6.7s. In backbone: RoI Selector 63ms (sum up), CPR\&CPS 4.2s (sum up).

\begin{table}[htbp]
    \caption{AGX Orin}
    \label{tab:X5}
    \centering
\begin{fullwidthtabular}{3}{ccc}
        \toprule
        Method & AP$_{50}$ & Latency(s) \\
        \midrule
        PAN \cite{fan2022speed} & 0.715 & 44.2 \\
        \midrule
        RT-DETR \cite{zhao2024detrs} & 0.612 & 22.5 \\
        \midrule
        DynamicDet \cite{Dynamicdet} & 0.639 & 17.3 \\
        \midrule
        DINO+SparseFormer \cite{SparseFormer} & \underline{0.780} & \underline{13.9} \\
        \midrule
        \rowcolor{RoyalBlue!20} DINO \cite{DINO}+Ours & \textbf{0.810} & \textbf{11.1} \\
        \bottomrule
        \end{fullwidthtabular}
\end{table}

\begin{table}[htbp]
    \caption{RTX 4090}
    \label{tab:X6}
    \centering
\begin{fullwidthtabular}{3}{ccc}
        \toprule
        Method & AP$_{50}$ & Latency(s) \\
        \midrule
        PAN \cite{fan2022speed} & 0.715 & 10.9 \\
        \midrule
        RT-DETR \cite{zhao2024detrs} & 0.612 & 4.2 \\
        \midrule
        DynamicDet \cite{Dynamicdet} & 0.639 & \underline{3.4} \\
        \midrule
        DINO+SparseFormer \cite{SparseFormer} & \underline{0.780} & 3.5 \\
        \midrule
        \rowcolor{RoyalBlue!20} DINO \cite{DINO}+Ours & \textbf{0.810} & \textbf{2.8} \\
        \bottomrule
        \end{fullwidthtabular}
\end{table}

\subsection{Detection on DOTA and VisDrone}
On DOTA \cite{DOTA}, we \textbf{improve mAP by 2\%} compared with SparseFormer \cite{SparseFormer} and \textbf{reduce 49\% FLOPs } compared with Efficientvit \cite{liu2023efficientvit}, shown in \Cref{tab:X7}. On VisDrone \cite{VisDrone}, we \textbf{improve AP$_{50}$ by 9\% and AR$_{100}$ by 11\%} compared with SparseFormer \cite{SparseFormer}, shown in \Cref{tab:X8}.

\begin{table}[htbp]
    \caption{DOTA}
    \label{tab:X7}
    \centering
\begin{fullwidthtabular}{3}{ccc}
        \toprule
        Method & mAP & GFLOPs \\
        \midrule
        DINO+Efficientvit \cite{liu2023efficientvit} & 74.39 & \underline{159.24} \\
        \midrule
        vHeat-T \cite{wang2025building} & 77.82 & 243.18 \\
        \midrule
        Swin-T \cite{Swin} & 77.63 & 221.32 \\
        \midrule
        DINO+SparseFormer \cite{SparseFormer} & \underline{77.94} & 174.31 \\
        \midrule
        \rowcolor{RoyalBlue!20} DINO \cite{DINO}+Ours & \textbf{79.52} & \textbf{89.63} \\
        \bottomrule
        \end{fullwidthtabular}
\end{table}

\begin{table}[htbp]
    \caption{VisDrone}
    \label{tab:X8}
    \centering
\begin{fullwidthtabular}{3}{ccc}
        \toprule
        Method & AP$_{50}$ & AR$_{100}$ \\
        \midrule
        DINO+Efficientvit \cite{liu2023efficientvit} & 0.473 & 0.338 \\
        \midrule
        vHeat-T \cite{wang2025building} & 0.512 & 0.385 \\
        \midrule
        Swin-T \cite{Swin} & 0.481 & 0.337 \\
        \midrule
        DINO+SparseFormer \cite{SparseFormer} & \underline{0.554} & \underline{0.408} \\
        \midrule
        \rowcolor{RoyalBlue!20} DINO \cite{DINO}+Ours & \textbf{0.602} & \textbf{0.452} \\
        \bottomrule
        \end{fullwidthtabular}
\end{table}

\subsection{Training stability}
We trained BiHashFormer with ten randomly drawn seeds and evaluated them on PANDA. Thanks to the binary nature of hash encoding, each pre-hash coordinate $z$ only needs to preserve its sign, rather than taking continuous integer values as in quantization. This substantially alleviates the impact of the STE on training stability. It is proved as all runs yield AP$_{50}\in[0.809,0.812]$.

\FloatBarrier
\subsection{Multi-category detection on COCO}
\label{sec:supp-coco}
We evaluate BiHashFormer on COCO~\cite{lin2014microsoft}, which contains 80 object categories, using Mask R-CNN with a $3\times$ training schedule. The Swin-T accuracy in \Cref{tab:X9} follows~\cite{Swin}. GFLOPs-O covers the backbone and neck at an input resolution of $1333\times800$, consistent with the computational scope used in the main paper. The average window selection ratio increases from 0.147 on PANDA to 0.30 on COCO, where foreground occupies a larger proportion of the image.

The Hamming dimension and prototype count serve different purposes. The 32-bit space encodes distinctions among categories, instances, and object parts, while the prototypes provide class-agnostic retrieval anchors for foreground structures. HashAttention uses binary codes to select key--value pairs and computes attention with their original continuous features. HashMiner uses the prototypes to retrieve missed foreground tokens, and the detection head predicts their categories. A prototype therefore need not correspond to a single object class.

\begin{table}[htbp]
    \caption{\textbf{Prototype-number ablation and comparison with Swin-T on COCO.} AP metrics are percentages. GFLOPs-O covers the backbone and neck; GFLOPs-CPS covers the compensation branch. Swin-T accuracy follows the $3\times$ Mask R-CNN results in~\cite{Swin}.}
    \label{tab:X9}
    \centering
    \setlength{\tabcolsep}{2.5pt}
    \begin{tabular*}{\linewidth}{@{\extracolsep{\fill}}lrrrrrr@{}}
        \toprule
        Backbone & \makecell{Prototype\\codes} & AP & $\mathrm{AP}_{50}$ & $\mathrm{AP}_{\mathrm{small}}$ & \makecell{GFLOPs\\O} & \makecell{GFLOPs\\CPS} \\
        \midrule
        Swin-T & -- & 46.0 & 68.1 & 31.2 & 138.34 & -- \\
        BiHashFormer & 1 & 47.0 & 70.2 & 32.4 & \textbf{86.96} & \textbf{5.01} \\
        BiHashFormer & 3 & \textbf{48.3} & \textbf{71.5} & \textbf{34.8} & 89.04 & 7.08 \\
        BiHashFormer & 5 & 47.2 & 70.8 & 33.1 & 96.83 & 14.88 \\
        \bottomrule
    \end{tabular*}
\end{table}

Three prototypes improve over Swin-T by 2.3 AP points, 3.4 $\mathrm{AP}_{50}$ points, and 3.6 $\mathrm{AP}_{\mathrm{small}}$ points with 35.6\% fewer GFLOPs. One prototype provides limited capacity to represent foreground modes. Increasing the count from three to five raises overall computation by 8.8\% without improving accuracy. Three prototypes give the best accuracy among the tested settings, consistent with the PANDA ablation. Together with the DOTA and VisDrone results, this evaluation extends the evidence from human-centric HRW images to aerial, drone-view, and conventional-resolution multi-category detection.

\FloatBarrier
\subsection{Comparison with efficient attention methods}
\label{sec:supp-efficient-attention}
We compare complete Linformer~\cite{wang2020linformer}, XAttention~\cite{xu2025xattention}, SparseFormer~\cite{SparseFormer}, and BiHashFormer model configurations on PANDA, each paired with the same DINO detection head~\cite{DINO}. \Cref{tab:X10} reports backbone/neck GFLOPs, inference latency, and detection accuracy for these full configurations.

Linformer projects the key and value sequences to a fixed rank before computing attention. This compression can mix weak small-object features with background features. HashAttention instead uses low-dimensional Hamming codes for query-specific selection and computes attention using the selected continuous features. XAttention estimates block importance through antidiagonal scoring. Its reported acceleration targets sequences of 8k--256k tokens, whereas a $21\times21$ window in our model contains 441 tokens. At this scale, block selection is coarser and its overhead is harder to amortize. BiHashFormer also uses HashMiner to recover target-related tokens outside the selected RoIs.

\begin{table}[htbp]
    \caption{\textbf{Comparison of complete model configurations on PANDA with the same DINO head.} GFLOPs-O covers the backbone and neck. Latency is in seconds, and $\mathrm{AP}_{50}$ is reported on the 0--1 scale.}
    \label{tab:X10}
    \centering
    \begin{tabular*}{\linewidth}{@{\extracolsep{\fill}}lrrr@{}}
        \toprule
        Configuration & GFLOPs-O $\downarrow$ & Latency (s) $\downarrow$ & $\mathrm{AP}_{50}$ $\uparrow$ \\
        \midrule
        Linformer + DINO & 57.48 & 3.15 & 0.792 \\
        XAttention + DINO & 64.93 & 3.25 & 0.799 \\
        SparseFormer + DINO & 75.71 & 3.5 & 0.780 \\
        BiHashFormer + DINO & \textbf{47.31} & \textbf{2.8} & \textbf{0.810} \\
        \bottomrule
    \end{tabular*}
\end{table}

BiHashFormer has the lowest computational cost and latency and the highest $\mathrm{AP}_{50}$ among these configurations. Relative to Linformer and XAttention, it improves $\mathrm{AP}_{50}$ by 1.8 and 1.1 percentage points, respectively.

\FloatBarrier
\subsection{Evaluation with DINOv3 pretraining}
\label{sec:supp-dinov3}
The main-paper PANDA results use training from scratch. To evaluate the method with strong pretrained features, we initialize both models in \Cref{tab:X11} from the same official DINOv3 ViT-S+/16 checkpoint~\cite{simeoni2025dinov3} and train for 36 epochs. ViT-S+/16 has 12 blocks and approximately 29 million parameters, close to the 12-block Swin-T/BiHashFormer configuration. Both models use identical input resolution, augmentation, optimizer, batch size, detector, losses, and evaluation settings.

We group the single-resolution ViT blocks as $[2,2,6,2]$ without downsampling. Within each block, the RoI Selector routes ranked spatial windows to HashAttention or HashMiner. The class token and four register tokens remain global and bypass this routing. We merge the branch outputs and restore the original token order. Both models load all shared pretrained components; the added routing, Hamming projection, prototype, and compensation modules are initialized from scratch.

\begin{table}[htbp]
    \caption{\textbf{PANDA accuracy and computation with identical DINOv3 initialization.} Both models use the same ViT-S+/16 checkpoint, detector, and 36-epoch protocol. F, B, and O denote foreground, background, and overall backbone/neck computation. AP metrics use the 0--1 scale.}
    \label{tab:X11}
    \centering
    \setlength{\tabcolsep}{2.5pt}
    \begin{tabular*}{\linewidth}{@{\extracolsep{\fill}}lrrrrr@{}}
        \toprule
        Backbone & \makecell{GFLOPs\\F} & \makecell{GFLOPs\\B} & \makecell{GFLOPs\\O} & $\mathrm{AP}_{50}$ & $\mathrm{AP}_{\mathrm{small}}$ \\
        \midrule
        \makecell[l]{DINOv3\\ViT-S+/16} & 23.61 & 448.71 & 472.32 & 0.768 & 0.411 \\
        \makecell[l]{BiHash-DINOv3\\ViT-S+/16} & 25.96 & \textbf{142.25} & \textbf{168.21} & \textbf{0.808} & \textbf{0.527} \\
        \bottomrule
    \end{tabular*}
\end{table}

\begin{table}[htbp]
    \caption{\textbf{PANDA wall-clock efficiency with DINOv3 pretraining.} Measurements use one RTX 4090.}
    \label{tab:X12}
    \centering
    \begin{tabular*}{\linewidth}{@{\extracolsep{\fill}}lrr@{}}
        \toprule
        Backbone & Latency (ms) $\downarrow$ & FPS $\uparrow$ \\
        \midrule
        DINOv3 ViT-S+/16 & 13,280 & 0.075 \\
        BiHash-DINOv3 ViT-S+/16 & \textbf{5,780} & \textbf{0.173} \\
        \bottomrule
    \end{tabular*}
\end{table}

BiHash-DINOv3 improves $\mathrm{AP}_{50}$ and $\mathrm{AP}_{\mathrm{small}}$ by 4.0 and 11.6 percentage points, respectively, while reducing overall GFLOPs by 64.4\%. As shown in \Cref{tab:X12}, latency decreases by 56.5\% and throughput increases by $2.31\times$. The small-object improvement supports the use of foreground-focused compression and missed-object compensation with pretrained features.

\FloatBarrier
\subsection{Training cost and convergence}
\label{sec:supp-training-cost}
We compare DINO + SparseFormer and DINO + BiHashFormer on PANDA using the same 36-epoch MMDetection protocol~\cite{chen2019mmdetection}. Training uses eight RTX 4090 GPUs with one image per GPU, for a total batch size of eight. \Cref{tab:X13} reports per-epoch and total training time, together with the epoch and elapsed time at which each model reaches $\mathrm{AP}_{50}=0.75$.

\begin{table}[htbp]
    \caption{\textbf{Training time and convergence on PANDA under the same 36-epoch protocol.} Both models use eight RTX 4090 GPUs and one image per GPU. AP metrics use the 0--1 scale.}
    \label{tab:X13}
    \centering
    \setlength{\tabcolsep}{2.5pt}
    \begin{tabular*}{\linewidth}{@{\extracolsep{\fill}}lrrrrr@{}}
        \toprule
        Method & \makecell{Time/epoch\\(min)} & \makecell{Total time\\(h)} & \makecell{Epoch to\\0.75 $\mathrm{AP}_{50}$} & \makecell{Time to 0.75\\$\mathrm{AP}_{50}$ (h)} & \makecell{Final\\$\mathrm{AP}_{50}$} \\
        \midrule
        \makecell[l]{DINO +\\SparseFormer} & 38.7 & 23.2 & 28 & 18.0 & 0.780 \\
        \makecell[l]{DINO +\\BiHashFormer} & \textbf{32.8} & \textbf{19.7} & \textbf{23} & \textbf{12.6} & \textbf{0.810} \\
        \bottomrule
    \end{tabular*}
\end{table}

BiHashFormer reduces total training time by 15.1\% and reaches $\mathrm{AP}_{50}=0.75$ five epochs earlier, taking 12.6\,h rather than 18.0\,h. It also reaches a higher final $\mathrm{AP}_{50}$. The inference measurements in \Cref{tab:X5,tab:X6} show corresponding practical speedups: 11.1\,s versus 13.9\,s on AGX Orin and 2.8\,s versus 3.5\,s on RTX 4090.

\FloatBarrier
\section{Code in PyTorch}
\label{sec:supp-code}
The following excerpts give the core operations of the public implementation at
\url{https://github.com/Purecoo00/BiHashFormer}. The accompanying
\texttt{pseudocode\_components.py} contains these functions and classes together.
Algorithms~\ref{alg:STE}--\ref{alg:Opt_PC} implement binary hashing, the Hamming
distance and the prototype objective. Algorithms~\ref{alg:HA} and~\ref{alg:HA-forward}
give the constructor, routing and forward pass of HashAttention.
Algorithm~\ref{alg:CPS} implements HashMiner, and Algorithm~\ref{alg:RoI}
constructs the per-image foreground targets and selection loss.

Hash codes are formed from each complete Q/K token vector before splitting it
into attention heads; the selected key indices are shared across heads.
HashAttention receives normalized RoI windows from the Hash Block. The
\texttt{valid} mask excludes padding, and \texttt{regions} prevents attention
across cyclic-shift boundaries. HashMiner reuses the compression branch's
normalization, K transform and key projector under \texttt{torch.no\_grad()};
these shared parameters are still trained by the compression branch.

During training, the Hamming distance uses an exact packed XOR/popcount forward
with a continuous-bit backward surrogate. A zero-forward affinity STE term in
the attention logits passes gradients through hard key routing without changing
the forward attention values. Window scoring uses an identity-forward STE gain
of $1+s-\operatorname{stopgrad}(s)$ on the selected window updates.
Window partitioning, scattering, window chunking and activation checkpointing
are handled by the Hash Block and are omitted here.

Foreground boxes and image dimensions use the same augmented image coordinate
system. The target tensor retains one ratio per image; squared errors and the
ratio penalty are computed per image before averaging over the batch and stages.

\begin{algorithm}[H]
\caption{Binary-code STE and Hamming distance (Equations (9)--(11))}
\label{alg:STE}
\begin{lstlisting}[language=python,firstnumber=1,firstline=13,lastline=50]
def binary_code_ste(z):
    """Hard 0/1 codes, including at zero; dz uses the paper's 1/2 STE."""
    hard = (z > 0).to(z.dtype)
    return hard + 0.5 * (z - z.detach())


def pairwise_hamming(a, b):
    """Packed 32-bit XOR/popcount forward, with a continuous STE backward.

    Leading batch dimensions are broadcast. The final two dimensions are
    (tokens, bits). Multiple words support the 64-bit ablation as well.
    """
    count = None
    for start in range(0, a.shape[-1], 32):
        size = min(32, a.shape[-1] - start)
        shifts = torch.arange(size, device=a.device, dtype=torch.int64)
        aa = (a[..., start : start + size].detach().long() << shifts).sum(-1)
        bb = (b[..., start : start + size].detach().long() << shifts).sum(-1)
        x = aa.unsqueeze(-1) ^ bb.unsqueeze(-2)
        x = x - ((x >> 1) & 0x55555555)
        x = (x & 0x33333333) + ((x >> 2) & 0x33333333)
        x = (x + (x >> 4)) & 0x0F0F0F0F
        x = x + (x >> 8)
        x = x + (x >> 16)
        word_count = x & 0x3F
        count = word_count if count is None else count + word_count
    hard = count.to(torch.float32)
    if torch.is_grad_enabled() and (a.requires_grad or b.requires_grad):
        # Exact on bits; this branch supplies the backward surrogate only.
        with torch.autocast(device_type=a.device.type, enabled=False):
            af, bf = a.float(), b.float()
            soft = (
                af.sum(-1, keepdim=True)
                + bf.sum(-1).unsqueeze(-2)
                - 2 * (af @ bf.transpose(-2, -1))
            )
        return hard + soft - soft.detach()
    return hard
\end{lstlisting}
\end{algorithm}

\begin{algorithm}[H]
\caption{Prototype initialization, losses and radius retrieval (Equations (13)--(16))}
\label{alg:Opt_PC}
\begin{lstlisting}[language=python,firstnumber=1,firstline=53,lastline=87]
class PrototypeCodes(nn.Module):
    def __init__(self, num_bits=32, num_prototypes=3, radius=4):
        super().__init__()
        if not 1 <= num_prototypes <= num_bits:
            raise ValueError("Require 1 <= num_prototypes <= num_bits")
        self.num_bits, self.radius = num_bits, radius
        # Start with separated codes so the prototypes cover different regions.
        latents = -torch.ones(num_prototypes, num_bits)
        for i in range(num_prototypes):
            latents[i, i * num_bits // num_prototypes : (i + 1) * num_bits // num_prototypes] = 1
        self.latents = nn.Parameter(latents)

    def bits(self):
        return binary_code_ste(self.latents)

    def loss(self, key_bits):
        centers = self.bits()
        if key_bits.numel() == 0:
            intra = self.latents.sum() * 0
        else:
            intra = (
                pairwise_hamming(key_bits.reshape(-1, self.num_bits), centers).min(-1).values.mean()
                / self.num_bits
            )
        if centers.shape[0] == 1:
            cross = centers.sum() * 0
        else:
            distances = pairwise_hamming(centers, centers)
            mask = ~torch.eye(len(centers), dtype=torch.bool, device=centers.device)
            cross = F.relu(2 * self.radius + 1 - distances[mask]).mean()
        return intra + cross

    def retrieve(self, key_bits):
        # A token is retained if it falls within the radius of any prototype.
        return pairwise_hamming(key_bits, self.bits().detach()).min(-1).values <= self.radius
\end{lstlisting}
\end{algorithm}

\begin{algorithm}[H]
\caption{HashAttention: projections and shared key routing}
\label{alg:HA}
\begin{lstlisting}[language=python,firstnumber=1,firstline=90,lastline=138]
class HashAttention(nn.Module):
    """Per-query nearest-half routing, attention in continuous feature space.

    The 32-bit routing code is formed from each complete token's Q/K vector;
    its selected key indices are shared by attention heads.
    """

    def __init__(
        self,
        dim,
        num_heads,
        window_size=21,
        num_bits=32,
        kv_ratio=0.5,
        num_prototypes=3,
        radius=4,
        attn_drop=0.0,
        query_chunk_size=64,
        routing_ste=True,
    ):
        super().__init__()
        if dim % num_heads or not 0 < kv_ratio <= 1:
            raise ValueError("Invalid head dimensions or KV ratio")
        self.dim, self.num_heads = dim, num_heads
        self.num_bits, self.kv_ratio = num_bits, kv_ratio
        self.scale = (dim // num_heads) ** -0.5
        self.query_chunk_size = query_chunk_size
        self.routing_ste = routing_ste
        self.qkv = nn.Linear(dim, 3 * dim)
        self.query_projector = nn.Linear(dim, num_bits, bias=False)
        self.key_projector = nn.Linear(dim, num_bits, bias=False)
        self.prototypes = PrototypeCodes(num_bits, num_prototypes, radius)
        self.proj = nn.Linear(dim, dim)
        self.dropout = nn.Dropout(attn_drop)
        w = window_size
        coords = torch.stack(
            torch.meshgrid(torch.arange(w), torch.arange(w), indexing="ij")
        ).flatten(1)
        relative = coords[:, :, None] - coords[:, None, :]
        index = (relative[0] + w - 1) * (2 * w - 1) + relative[1] + w - 1
        self.register_buffer("relative_index", index, persistent=False)
        self.relative_bias = nn.Parameter(torch.zeros((2 * w - 1) ** 2, num_heads))

    def route(self, q_bits, k_bits, valid, regions):
        distance = pairwise_hamming(q_bits, k_bits)
        allowed = valid[:, None, :] & (regions[:, :, None] == regions[:, None, :])
        distance = distance.masked_fill(~allowed, self.num_bits + 1e4)
        topk = max(1, int(q_bits.shape[-2] * self.kv_ratio))
        return distance.topk(topk, dim=-1, largest=False), allowed
\end{lstlisting}
\end{algorithm}

\begin{algorithm}[H]
\caption{HashAttention: attention on the routed key--value pairs}
\label{alg:HA-forward}
\begin{lstlisting}[language=python,firstnumber=1,firstline=140,lastline=178,gobble=4]
    def forward(self, x, valid, regions):
        b, n, c = x.shape
        q, k, v = self.qkv(x).chunk(3, -1)
        qb = binary_code_ste(self.query_projector(q))
        kb = binary_code_ste(self.key_projector(k))
        (selected_distances, indices), allowed = self.route(qb, kb, valid, regions)
        qh, kh, vh = [
            t.reshape(b, n, self.num_heads, c // self.num_heads).transpose(1, 2) for t in (q, k, v)
        ]
        bias = self.relative_bias[self.relative_index].permute(2, 0, 1)
        batch = torch.arange(b, device=x.device)[:, None, None, None]
        head = torch.arange(self.num_heads, device=x.device)[None, :, None, None]
        chunks = []
        for start in range(0, n, self.query_chunk_size):
            end = min(start + self.query_chunk_size, n)
            idx = indices[:, start:end]
            selected_k = kh[batch, head, idx[:, None]]
            selected_v = vh[batch, head, idx[:, None]]
            logits = (qh[:, :, start:end, None] * selected_k).sum(-1) * self.scale
            local_bias = bias[:, start:end].unsqueeze(0).expand(b, -1, -1, -1)
            logits = logits + local_bias.gather(-1, idx[:, None].expand(-1, self.num_heads, -1, -1))
            if self.training and self.routing_ste:
                # Top-k indices have no gradient. This surrogate trains the hash
                # projectors without changing the attention values.
                affinity = -selected_distances[:, start:end] / self.num_bits
                logits = logits + (affinity - affinity.detach())[:, None]
            permitted = allowed[:, start:end].gather(-1, idx)
            logits = logits.masked_fill(~permitted[:, None], torch.finfo(logits.dtype).min)
            weights = self.dropout(logits.softmax(-1))
            chunks.append((weights[..., None] * selected_v).sum(-2))
        out = torch.cat(chunks, dim=2).transpose(1, 2).reshape(b, n, c)
        out = self.proj(out) * valid[..., None]
        if self.training:
            seen = torch.zeros(b, n, dtype=torch.bool, device=x.device)
            seen.scatter_(1, indices.reshape(b, -1), True)
            loss = self.prototypes.loss(kb[seen & valid])
        else:
            loss = x.new_zeros(())
        return out, loss
\end{lstlisting}
\end{algorithm}

\begin{algorithm}[H]
\caption{HashMiner: retrieval and attention within each non-RoI window}
\label{alg:CPS}
\begin{lstlisting}[language=python,firstnumber=1,firstline=181,lastline=222]
class HashMiner(nn.Module):
    """Radius-based non-RoI retrieval followed by attention on retained tokens."""

    def __init__(self, dim, num_heads, mlp_ratio=4, attn_drop=0.0):
        super().__init__()
        self.norm1, self.norm2 = nn.LayerNorm(dim), nn.LayerNorm(dim)
        self.attn = nn.MultiheadAttention(dim, num_heads, dropout=attn_drop, batch_first=True)
        self.ffn = nn.Sequential(
            nn.Linear(dim, int(dim * mlp_ratio)), nn.GELU(), nn.Linear(int(dim * mlp_ratio), dim)
        )

    def forward(self, x, valid, regions, compression, compression_norm):
        with torch.no_grad():
            normalized = compression_norm(x)
            c = compression.dim
            # Reuse the compression key transform/projector without updating
            # them from the compensation branch. Compression still trains them.
            keys = F.linear(
                normalized, compression.qkv.weight[c : 2 * c], compression.qkv.bias[c : 2 * c]
            )
            bits = binary_code_ste(compression.key_projector(keys))
            selected = compression.prototypes.retrieve(bits) & valid
        out = x.clone()
        # Group equal-length selections to avoid padding attention sequences.
        counts = selected.sum(-1)
        for count in counts.unique().tolist():
            if count == 0:
                continue
            windows = (counts == count).nonzero(as_tuple=True)[0]
            positions = selected[windows].nonzero(as_tuple=False)[:, 1].reshape(-1, count)
            tokens = x[windows[:, None], positions]
            labels = regions[windows[:, None], positions]
            blocked = labels[:, :, None] != labels[:, None, :]
            blocked = (
                blocked[:, None].expand(-1, self.attn.num_heads, -1, -1).reshape(-1, count, count)
            )
            normed = self.norm1(tokens)
            update, _ = self.attn(normed, normed, normed, attn_mask=blocked, need_weights=False)
            tokens = tokens + update
            tokens = tokens + self.ffn(self.norm2(tokens))
            out[windows[:, None], positions] = tokens
        return out, selected
\end{lstlisting}
\end{algorithm}

\begin{algorithm}[H]
\caption{Per-image foreground targets and selection loss (Equations (5) and (8))}
\label{alg:RoI}
\begin{lstlisting}[language=python,firstnumber=1,firstline=225,lastline=256]
def foreground_ratio(boxes, image_shape):
    """Exact clipped union area, rather than the sum of overlapping boxes."""
    height, width = image_shape[:2]
    if height <= 0 or width <= 0:
        raise ValueError("Image dimensions must be positive")
    if hasattr(boxes, "tensor"):
        boxes = boxes.tensor
    # Overlapping boxes should contribute to foreground coverage only once.
    rectangles = []
    for x1, y1, x2, y2 in boxes.detach().cpu().tolist():
        x1, x2 = max(0.0, x1), min(float(width), x2)
        y1, y2 = max(0.0, y1), min(float(height), y2)
        if x2 > x1 and y2 > y1:
            rectangles.append(box(x1, y1, x2, y2))
    return unary_union(rectangles).area / (height * width) if rectangles else 0.0


def foreground_targets(batch_inputs, batch_data_samples):
    """Keep one foreground target for each image in the batch."""
    return batch_inputs.new_tensor(
        [
            foreground_ratio(sample.gt_instances.bboxes, sample.img_shape)
            for sample in batch_data_samples
        ]
    )


def selection_loss(stage_ratios, targets, gamma=0.1):
    """Average per-image losses, then average over stages."""
    return torch.stack(
        [((ratio - targets) ** 2 + gamma * ratio).mean() for ratio in stage_ratios]
    ).mean()
\end{lstlisting}
\end{algorithm}

\clearpage
\section{Enlarged figure details}
\label{sec:figure-details}
\begin{figure}[htbp]
    \centering
    \includegraphics[width=\linewidth,trim=390pt 0pt 0pt 222pt,clip]{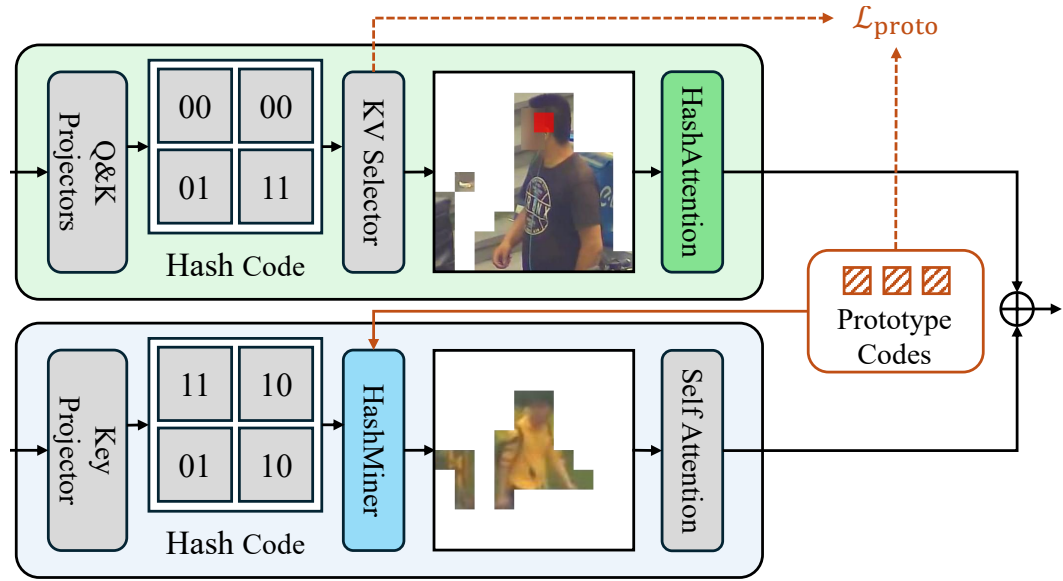}
    \caption{Enlarged detail from the architecture overview in Figure~2. The vector crop shows the HashAttention and HashMiner pathways, prototype codes, and their connections.}
    \label{fig:architecture-detail}
\end{figure}

\begin{figure}[htbp]
    \centering
    \includegraphics[width=0.9\linewidth]{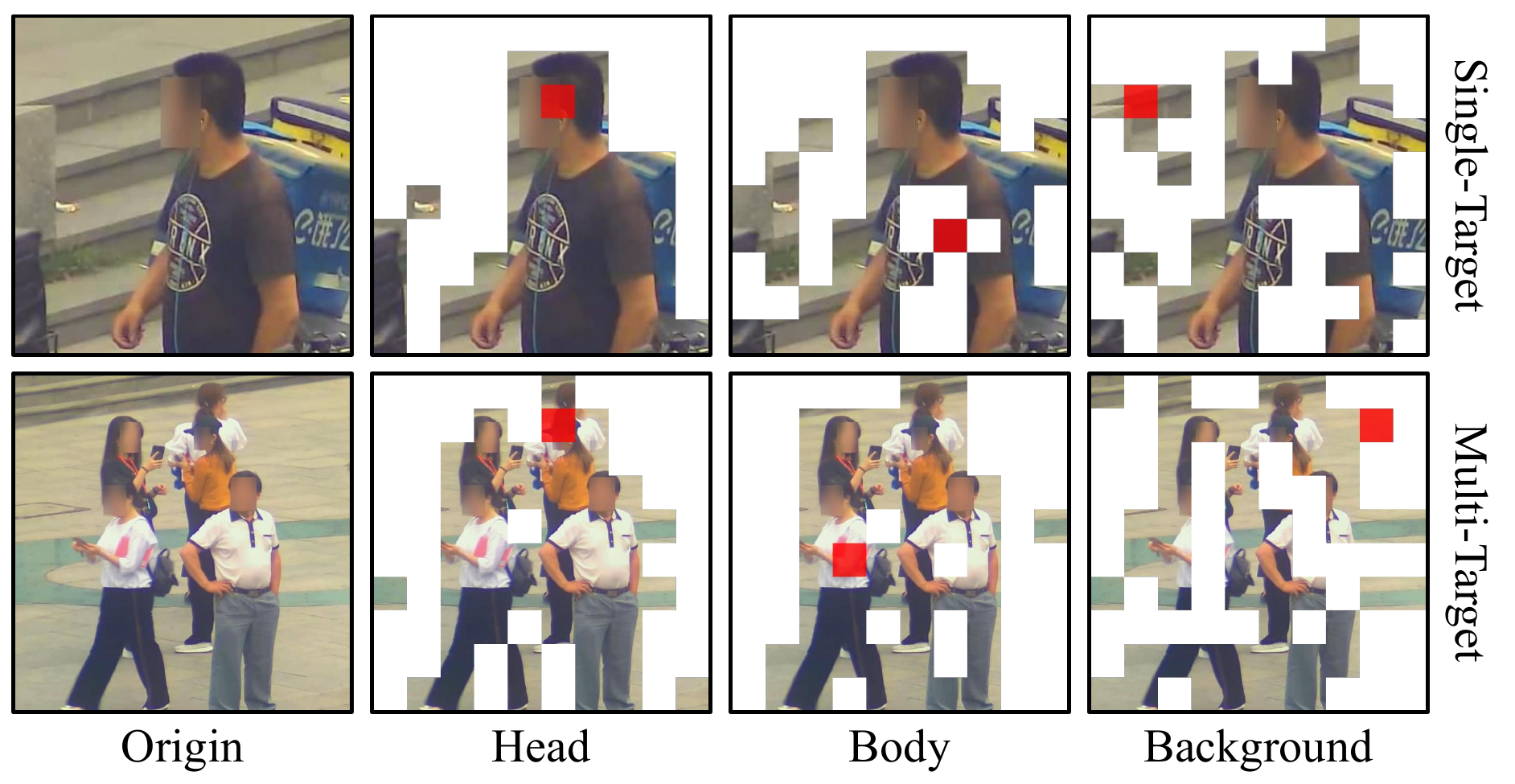}
    \caption{Enlarged view of the query--key visualization in Figure~3. The original figure content is unchanged.}
    \label{fig:attention-detail}
\end{figure}

Figures~\ref{fig:architecture-detail} and~\ref{fig:attention-detail} provide enlarged vector views of the main-paper figures for easier inspection.

\FloatBarrier
\newpage
\bibliographystyle{named}
\bibliography{neurips_2026}